\documentclass{article}

\PassOptionsToPackage{numbers, compress}{natbib}

\usepackage[preprint]{neurips_2026}

\usepackage[utf8]{inputenc} 
\usepackage[T1]{fontenc}    
\usepackage{hyperref}       
\usepackage{url}            
\usepackage{booktabs}       
\usepackage{amsfonts}       
\usepackage{nicefrac}       
\usepackage{microtype}      
\usepackage{xcolor}         
\usepackage{enumitem}
\usepackage{amsmath}
\usepackage{multirow}
\usepackage{subcaption}
\usepackage{wrapfig}
\usepackage{pifont}
\usepackage{marvosym}
\usepackage{tabularx}
\usepackage{graphicx}
\usepackage{float}
\usepackage{makecell}
\usepackage[table]{xcolor}
\usepackage{algorithm}
\usepackage{algpseudocode}
\usepackage{titletoc}
\definecolor{bg_gray}{gray}{0.92}

\title{Perceive, Route and Modulate: Dynamic Pattern Recalibration for Time Series Forecasting}

\author{
 Siru Zhong\textsuperscript{1} \thanks{Work done during Siru Zhong's stay at Tsinghua University as a visiting PhD student.} 
 \qquad Zhao Meng\textsuperscript{2}   \qquad Haohuan Fu\textsuperscript{2} \\ \bf Haoyang Li\textsuperscript{3} \qquad Qingsong Wen\textsuperscript{3} \qquad Yuxuan Liang\textsuperscript{1 \Letter} \\[1mm]
  \mdseries
  \textsuperscript{1} The Hong Kong University of Science and Technology (Guangzhou) \\
  \textsuperscript{2} Tsinghua University
  \textsuperscript{3} Squirrel Ai Learning
}

\begin{document}

\maketitle

\begin{abstract}
Local temporal patterns in real-world time series continuously shift, rendering globally shared transformations suboptimal. Current deep forecasting models, despite their scale and complexity, rely on fixed weight matrices applied uniformly to all temporal tokens. This creates a \textit{static pattern response}: models settle into a compromised average, unable to adapt to changing local dynamics. We introduce \textbf{Dynamic Pattern Recalibration} (DPR), a backbone-agnostic mechanism that resolves this via token-level recalibration. Through a lightweight ``Perceive-Route-Modulate'' pipeline, DPR computes a soft-routing distribution over a learned basis of adaptive response patterns, generating a time-aware modulation vector that recalibrates hidden states via a residual Hadamard product. As a backbone-agnostic adapter, DPR enhances forecasting across diverse architectures with minimal overhead, confirming it addresses a general bottleneck. As a minimalist standalone model, DPRNet achieves competitive performance across 12 benchmarks, validating dynamic recalibration against macroscopic parameter scaling. 
\end{abstract}

\section{Introduction}

Time series forecasting is central to decision-making in systems characterized by continuous volatility, ranging from high-frequency financial markets and energy grids to climate prediction, epidemiology, and system diagnostics \citep{idrees2019prediction,karevan2020transductive,deb2017review,zheng2020traffic,ruan2025cross,zhang2024predicting,liu2025towards,zou2025fine}. A fundamental challenge across these domains is that real-world systems exhibit complex local dynamics: alternating among predictable stable trends, periodic oscillations, and sudden anomalies \citep{hamilton1990analysis,liu2023anomaly,zhong2026dropoutts}. In essence, local temporal patterns continuously change, rendering globally shared feature transformations suboptimal: the same mapping cannot simultaneously serve stable periods and volatile shocks. Robust forecasting therefore requires models that dynamically adjust their feature sensitivity to match changing local dynamics in real time \citep{han2022dynamic, chen2020dynamic}.

Despite rapid advancements in deep forecasting architectures \citep{nie2023patchtst, liu2024itransformer} and the recent push toward massive horizontal scaling via dense foundation models \citep{das2024decoder, ansarichronos} or sparse Mixture-of-Experts (MoE) \citep{shi2025timemoe, ma2025timeexpert}, a critical microscopic constraint remains unresolved. Standard architectures apply feature transformations (e.g., feed-forward networks, linear projections) using weight matrices $\Theta$ that are frozen after training and shared globally across all temporal tokens. While attention mechanisms can dynamically aggregate context, the subsequent transformations that shape the actual hidden representations remain rigid. This enforces a \textit{static pattern response}: rigid transformations cannot smoothly alternate between the conservative mapping required for stable periods and the high-sensitivity amplification necessary during volatile shocks, ultimately forcing the model into a compromised average representation. Existing statistical mitigations, such as RevIN \citep{kim2021reversible}, operate purely as sequence-level preprocessing and fail to alter the backbone's token-level mapping limitations.

\begin{figure}[t!]
  \centering
  \includegraphics[width=\linewidth]{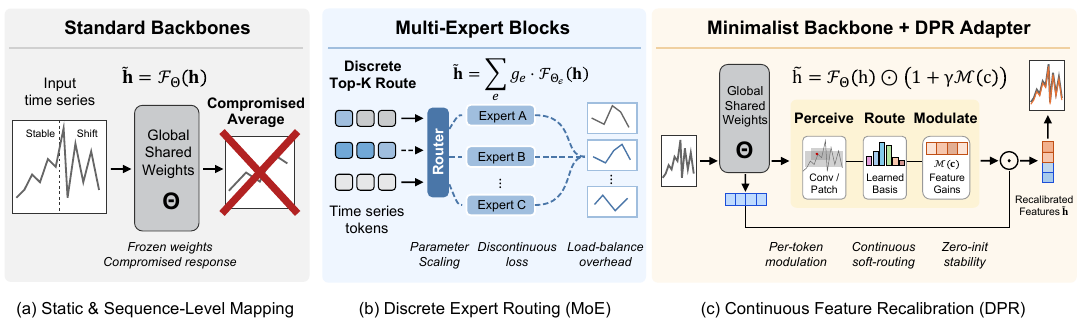}
  \caption{Comparison of forecasting paradigms. \textbf{(a)} Standard backbones: fixed mapping compromises across local dynamics. \textbf{(b)} MoE: discrete expert routing scales parameters and requires load balancing. \textbf{(c)} DPR: dynamic pattern recalibration via a lightweight Perceive-Route-Modulate mechanism.}
  \vspace{-1em}
  \label{fig:dpr_intro}
\end{figure}

To solve this, we propose \textbf{Dynamic Pattern Recalibration} (DPR), a general mechanism that decouples global temporal mapping from local dynamic adaptation. Unlike model-specific architectural innovations, DPR is a backbone-agnostic adapter: it replaces rigid, globally shared feature transformations with continuous token-level pattern recalibration, and applies universally across attention, convolution, MLP, and GNN backbones. DPR operates on an orthogonal axis to attention and MoE: attention determines which tokens to aggregate, MoE determines which parameters to activate, and DPR determines how to modulate feature dimensions per token. As illustrated in Figure~\ref{fig:dpr_intro}, standard backbones enforce static, token-level mappings that compromise across varying local dynamics (a), while MoE architectures introduce discrete token routing with substantial overhead and operate agnostically to temporal locality (b). DPR instead resolves the static pattern response via a lightweight \textit{Perceive-Route-Modulate} pipeline without modifying the host backbone's architecture (c).

Concretely, DPR senses dynamics through multi-scale convolutions or patches (Perceive), computes a soft-routing distribution over a learned, orthogonally regularized basis of adaptive response patterns (Route), and applies token-level pattern recalibration via a residual Hadamard product (Modulate). We demonstrate generality through two instantiations: DPRNet, a minimalist patch-based MLP validating dynamic recalibration against macroscopic parameter scaling, and a backbone-agnostic adapter integrated into seven mainstream forecasting backbones. Our contributions are:

\begin{itemize}[leftmargin=*]
    \item \textbf{Dynamic Pattern Recalibration.} We identify the \textit{static pattern response} in standard forecasting backbones and propose DPR for token-level continuous pattern recalibration.
    \item \textbf{Backbone-Agnostic Enhancement.} DPR consistently improves forecasting across seven mainstream architectures (attention, convolution, MLP, GNN) with minimal parameter overhead.
    \item \textbf{A Minimalist Instantiation (DPRNet).} We instantiate DPR in a minimalist patch-based MLP, achieving competitive performance across twelve benchmarks.
\end{itemize}
\section{Preliminary}
\label{sec:prelim_related}

\subsection{Problem Formulation}
\label{subsec:problem_statement}

In multivariate time series forecasting, we aim to predict a future horizon $\mathcal{Y} \in \mathbb{R}^{S \times C}$ given a lookback window $\mathcal{X} \in \mathbb{R}^{T \times C}$. Deep forecasters encode $\mathcal{X}$ into hidden states $\mathbf{H} \in \mathbb{R}^{B \times L \times d}$, where each layer applies a base transformation $\mathcal{F}_{\Theta}$ (FFN, linear projection, or convolution) to every temporal position:
\begin{equation}
\mathbf{h}_\ell^{(l)} = \mathcal{F}_{\Theta}\bigl(\mathbf{h}_\ell^{(l-1)}\bigr), \qquad
\tilde{\mathbf{h}}_\ell^{(l)} = \mathcal{F}_{\Theta}\bigl(\mathbf{h}_\ell^{(l-1)}\bigr) \odot \mathcal{M}(\mathbf{c}_\ell)
\label{eq:dynamic_recalibration}
\end{equation}
The left gives the standard static mapping with globally frozen $\Theta$, which enforces a compromised average: unable to simultaneously serve stable trends and volatile shifts. We term this the \textit{static pattern response}. The right introduces dynamic recalibration via local modulation $\mathcal{M}(\mathbf{c}_\ell)$, where $\mathbf{c}_\ell$ captures the local temporal context at position $\ell$, adjusting feature gains per token without modifying the backbone's weights ($\odot$: Hadamard product). 

Intuitively, $\mathcal{M}(\mathbf{c}_\ell)$ learns *when* to amplify or attenuate each feature dimension locally, while $\Theta$ retains the global mapping; this decomposition decouples global representation from local adaptation.

\subsection{Related Work}
\label{subsec:related_work}

\textbf{Deep Time Series Forecasting Architectures.} 
Recent work on deep forecasting has pursued two main directions. The first designs structural priors, spanning attention \citep{vaswani2017attention, zhou2021informer, wu2021autoformer, zhang2023crossformer, nie2023patchtst, liu2024itransformer}, linear mappings \citep{zeng2023are}, 2D convolutions \citep{wu2023timesnet}, multi-scale mixing \citep{wang2024timemixer,Murad_2025}, and GNN filtering \citep{hu2025timefilter}. The second scales via dense foundation models \citep{das2024decoder, ansarichronos, woo2024unified} or sparse MoE with discrete routing \citep{shazeer2017sparselygated, fedus2022switch, puigcerver2024from, shi2025timemoe, ma2025timeexpert}. While standard backbones apply static token-level transformations, MoE introduces input-dependent routing but relies on discrete expert selection, which creates discontinuous loss landscapes and demands careful load-balancing, limiting smooth tracking of pattern transitions. DPR addresses this by replacing discrete routing with a continuous soft combination of learned adaptive response patterns. Structurally, DPR differs from both MoE and attention. MoE routes whole tokens to discrete experts. Attention aggregates across temporal positions. DPR performs \textit{intra-token} recalibration: modulating individual feature dimensions per token, conditioned on local context, without load-balancing overhead or cross-position mixing.

\textbf{Conditional Feature Recalibration.}
Adaptive recalibration has been widely adopted in computer vision, with mechanisms like FiLM \citep{perez2018film} generating context-dependent affine parameters, SE-Net \citep{hu2018squeeze} and CBAM \citep{woo2018cbam} recalibrating channel importance, and CondConv \citep{yang2019condconv} dynamically generating input-dependent kernels. However, these techniques typically rely on \textit{global spatial pooling} to derive their recalibration signals. Applying this design directly to time series with diverse local dynamics is fundamentally flawed: aggregating global sequential statistics effectively averages out the very pattern transitions and transient shocks the model needs to detect. Therefore, effective temporal pattern recalibration must be inferred exclusively from \textit{local} sequential context to provide fine-grained, token-level adaptation. Within the time series domain, existing mitigations operate at the wrong granularity to solve the static pattern response. Stationarization techniques like RevIN \citep{kim2021reversible} perform sequence-level affine scaling, which merely normalizes the input distribution but does not alter the backbone's static feature mapping. Non-stationary Transformers~\citep{liu2022non} and Koopa~\citep{liu2023koopa} target non-stationarity at the architecture level, yet their downstream feed-forward networks remain uncalibrated. In contrast, DPR grounds its recalibration in local temporal context alone. Each token combines a globally learned basis of adaptive response patterns into a recalibration vector that dynamically recalibrates the backbone's feature response, yielding fine-grained adaptation to local dynamics at negligible additional parameter cost.

\section{Methodology}
\label{sec:method}

\begin{figure}[t]
  \centering
  \includegraphics[width=\linewidth]{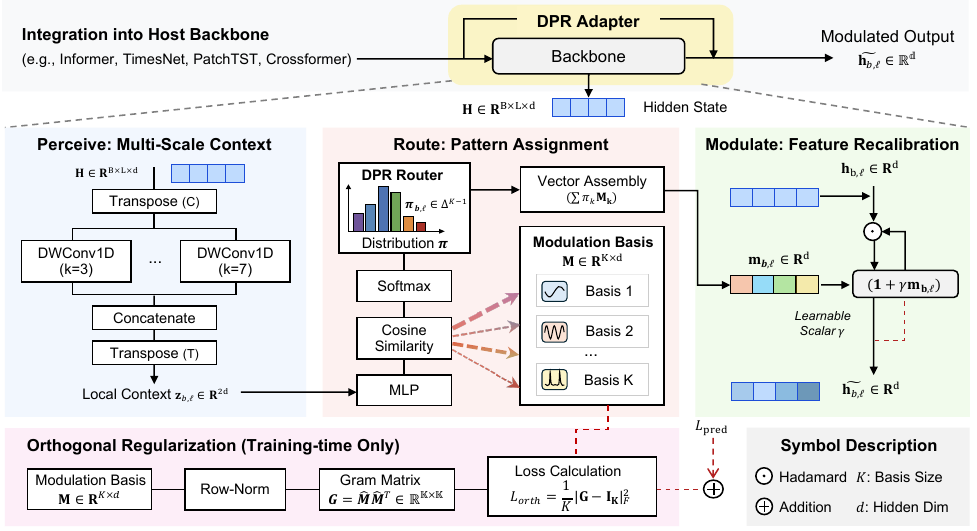}
  \caption{Overview of DPR. (Top) Backbone-agnostic adapter. (Middle) Perceive-Route-Modulate: multi-scale convolutions extract local context; soft-routing over $K$ learnable basis vectors; Hadamard modulation recalibrates features. (Bottom) Orthogonal regularization for basis diversity.}
  \label{fig:DPR-framework}
\end{figure}

To instantiate dynamic pattern recalibration (Eq.~\ref{eq:dynamic_recalibration}), we elaborate \textbf{Dynamic Pattern Recalibration} (DPR). Figure~\ref{fig:DPR-framework} provides an overview of the architecture: a backbone-agnostic adapter (top) encapsulates the three-stage Perceive-Route-Modulate pipeline (middle), which computes token-level modulation vectors from local context, regularized by an orthogonal basis constraint (bottom). DPR thus decomposes temporal modeling into two orthogonal responsibilities: a minimalist backbone performs base global temporal mapping, while DPR recalibrates features conditioned on local context.

\subsection{The Minimalist DPRNet Architecture}
\label{subsec:dpr_net_arch}

To isolate and validate the efficacy of pattern recalibration over macroscopic parameter scaling, DPRNet deliberately employs a simplified, Transformer-free backbone.

Given a historical sequence $\mathcal{X} \in \mathbb{R}^{T \times C}$, we adopt Channel Independence (CI) \citep{nie2023patchtst}, treating each variable as an independent 1D sequence and applying Reversible Normalization (RevIN) \citep{kim2021reversible} to mitigate distributional shifts. To preserve local semantic continuity and compress the temporal horizon, the sequence is segmented into $L$ overlapping patches (length $P$, stride $S$) and linearly projected to a hidden dimension $d$, yielding the initial representation $\mathbf{H}^{(0)} \in \mathbb{R}^{B \times L \times d}$.

The core of the network is a stack of DPR Blocks. To ensure optimization stability, we utilize residual connections. At layer $l$, the hidden state is first processed by a static base mapping:
\begin{equation}
\mathbf{Z}^{(l)} = \mathbf{H}^{(l-1)} + \mathcal{F}_{\mathrm{MLP}}\left( \mathrm{LayerNorm}(\mathbf{H}^{(l-1)}) \right)
\end{equation}
where $\mathcal{F}_{\mathrm{MLP}}$ is a standard multi-layer perceptron (GELU activations). Because $\mathcal{F}_{\mathrm{MLP}}$ relies on static, globally shared parameters, it inherently suffers from the static pattern response. We resolve this by immediately cascading it with our DPR adapter, which performs token-level recalibration:
\begin{equation}
\mathbf{H}^{(l)} = \mathbf{Z}^{(l)} + \mathrm{DPR}\left( \mathrm{LayerNorm}(\mathbf{Z}^{(l)}) \right)
\end{equation}
After $N_{\mathrm{blocks}}$ layers, the final hidden sequence $\mathbf{H}^{(N_{\mathrm{blocks}})}$ is temporally flattened and decoded by a linear head to yield the prediction $\mathcal{Y} \in \mathbb{R}^{S \times C}$, followed by RevIN denormalization.

\subsection{Dynamic Pattern Recalibration (DPR)}
\label{subsec:dpr_module}

The DPR mechanism is a lightweight component designed to extract time-aware modulation vectors from a learned basis of adaptive response patterns. For a hidden state $\mathbf{h}_{b,\ell} \in \mathbb{R}^d$ at temporal position $\ell \in \{1, \dots, L\}$, DPR modulates the feature response through three stages:

\textbf{Perceive: Multi-Scale Context.} To dynamically detect localized behaviors (e.g. abrupt spikes or meso-scale pattern transitions), DPR extracts context features via multi-scale depthwise 1D convolutions~\citep{bai2018empirical}. By transposing the input to a channel-first layout $\mathbf{H}^{\top} \in \mathbb{R}^{B \times d \times L}$, we apply two parallel depthwise convolutions with diverse receptive fields (e.g., kernels $k_1=3, k_2=7$):
\begin{equation}
    \mathbf{Z} = \text{Concat}\left( \text{DWConv}_{k_1}(\mathbf{H}^{\top}), \text{DWConv}_{k_2}(\mathbf{H}^{\top}) \right)^{\top}
\label{eq:local_perception}
\end{equation}
yielding the perceived context $\mathbf{Z} \in \mathbb{R}^{B \times L \times 2d}$. Crucially, depthwise operations prevent cross-channel information leakage, maintaining the strict CI assumption. For patch-based backbones where individual tokens inherently encapsulate sufficient local span, this operation smoothly reduces to a pointwise convolution ($k=1$) to prevent temporal over-smoothing.

\textbf{Route: Pattern Assignment.} The perceived context $\mathbf{z}_{b,\ell} \in \mathbb{R}^{2d}$ is compressed via an MLP to a compact context query $\mathbf{c}_{b,\ell} = \mathrm{GELU}(\mathbf{W}_1 \mathbf{z}_{b,\ell} + \mathbf{b}_1) \in \mathbb{R}^{d_c}$, where $d_c = \max(16, \lfloor d/4 \rfloor)$. 

We introduce $K$ learnable pattern centroids $\mathbf{E} \in \mathbb{R}^{K \times d_c}$ acting as ``keys'' to anchor distinct response patterns. The soft-routing probability vector $\boldsymbol{\pi}_{b,\ell} \in \Delta^{K-1}$ is computed via scaled cosine similarity:
\begin{equation}
\pi_{b,\ell,k} = \mathrm{softmax} \left( \tau \cdot \langle \hat{\mathbf{c}}_{b,\ell}, \hat{\mathbf{E}}_k \rangle \right)
\end{equation}
where $\hat{\mathbf{c}}$ and $\hat{\mathbf{E}}$ are $L_2$-normalized vectors, and $\tau$ is a learnable temperature scale. This normalization prevents magnitude discrepancies from dominating pattern assignment, ensuring stable gradients.

\textbf{Modulate: Feature Recalibration.} DPR maps the routing probabilities to a learned basis of adaptive response patterns $\mathbf{M} \in \mathbb{R}^{K \times d}$, generating a modulation vector via convex combination and recalibrating the hidden state through a Hadamard modulation:
\begin{equation}
\mathbf{m}_{b,\ell} = \sum_{k=1}^{K} \pi_{b,\ell,k} \mathbf{M}_{k,:}, \qquad
\tilde{\mathbf{h}}_{b,\ell} = \mathbf{h}_{b,\ell} \odot \bigl(\mathbf{1} + \gamma \mathbf{m}_{b,\ell}\bigr)
\label{eq:residual_modulation}
\end{equation}
where $\gamma$ is a learnable scalar initialized to $0$. Inspired by \citep{he2016deep}, zero-initialization makes DPR begin as an \textit{identity mapping}. The residual prior preserves the backbone's representation during early training, preventing destabilization while gradually introducing recalibration. A theoretical analysis of this inductive bias is provided in Appendix~\ref{sec:appendix_inductive_bias}.

\textbf{Orthogonal Regularization.} To preclude mode collapse (where the routing distribution degenerates into a uniform mixture or redundant bases), we impose a hard structural constraint on the modulation matrix $\mathbf{M}$ toward orthogonality. Let $\hat{\mathbf{M}}$ denote the row-wise $L_2$-normalized basis and $\mathbf{G} = \hat{\mathbf{M}}\hat{\mathbf{M}}^\top$ its associated Gram matrix. The orthogonal penalty is defined as:
\begin{equation}
\mathcal{L}_{\mathrm{orth}} = \frac{1}{K} \left\| \mathbf{G} - \mathbf{I}_K \right\|_F^2
\end{equation}
The total objective is $\mathcal{L}_{\mathrm{total}} = \mathcal{L}_{\mathrm{pred}} + \lambda_{\mathrm{orth}} \mathcal{L}_{\mathrm{orth}}$. By enforcing orthogonality among basis vectors, $\mathcal{L}_{\mathrm{orth}}$ yields sharp routing under phase shifts and parsimonious representations during stable periods.

\subsection{Complexity Analysis and Efficiency}
\label{subsec:complexity}

\textbf{Parameter Scalability.} The adapter adds three lightweight components: depthwise convolutions ($d \cdot k$), an MLP projection ($d \cdot d_c$), and the modulation basis ($K \cdot d$). With $K \ll d$ and $d_c \le d/4$, the dominant cost is a single $d \times d_c$ matrix, negligible next to any backbone.

\textbf{Computational Efficiency.} Unlike sparse MoE, which introduces discrete routing discontinuities and complex load-balancing, DPR's soft-routing is deterministic and trivially parallelizable. The Hadamard modulation adds only $\mathcal{O}(L \cdot d)$ operations, preserving linear temporal complexity.

\subsection{Discussion: Relation to Attention and MoE}
\label{subsec:discussion}

DPR is backbone-agnostic: it resolves the static pattern response by decoupling global temporal mapping from local pattern recalibration. Consistent gains across diverse backbones (Section~\ref{sec:universal_enhancement}) confirm this addresses a general bottleneck, with integration details in Appendix~\ref{sec:appendix_integration}.

\vspace{-1em}
\begin{table}[h!]
    \centering
    \caption{\textbf{Feature transformation paradigms.} From static weights (Level~0) to local context-aware recalibration (Level~2). DPR (Level~2): per-dimension modulation without load-balancing. $^\dagger$Qualitative. $^\ddagger$Requires auxiliary load-balancing~\citep{fedus2022switch}.}
    \label{tab:paradigm}
    \resizebox{\textwidth}{!}{
    \begin{tabular}{l c c c c c c}
    \toprule
    \textbf{Paradigm (Example)} & \textbf{Level} & \textbf{Feature Transform} & \textbf{Conditioning} & \textbf{Routing} & \textbf{Optimization} & \textbf{Overhead$^\dagger$} \\
    \midrule
    Static Mapping (Attention, RevIN) 
    & 0 & Static & Global / Sequence-level & N/A & Stable & None / Negligible \\
    
    Expert Routing (MoE) 
    & 1 & Dynamic (expert mixing) & Token-level & Discrete Top-$K$ & Unstable$^\ddagger$ & High \\
    
    \rowcolor{gray!10}
    \textbf{Pattern Recalibration (DPR)} 
    & \textbf{2} & \textbf{Dynamic (recalibration)} & \textbf{Local $\to$ Token-level} & \textbf{Soft} & \textbf{Stable} & \textbf{Minimal} \\
    \bottomrule
    \end{tabular}
    }
\end{table}

Table~\ref{tab:paradigm} organizes forecasting models along a continuum of increasing adaptation granularity:

\begin{enumerate}[leftmargin=*]
    \item \textbf{Static Mapping.} Standard backbones (PatchTST, iTransformer) apply globally shared frozen weights to every token, producing the \textit{static pattern response} defined in Section~\ref{subsec:problem_statement}. Sequence-level conditioning (RevIN)~\citep{kim2021reversible} injects global statistics but does not alter the per-layer transformation, remaining insufficient for token-level pattern recalibration.
    \item \textbf{Expert Routing (MoE).} Sparse MoE~\citep{shazeer2017sparselygated} routes tokens to discrete expert subsets via discrete Top-$K$, achieving token-level adaptation at the cost of a discontinuous loss landscape, auxiliary load-balancing, and routing decisions made without local context.
    \item \textbf{Pattern Recalibration (DPR).} DPR replaces discrete expert selection with soft combination of an orthogonal basis of adaptive response patterns, yielding per-dimension, token-level modulation from local context; intra-token recalibration without cross-position mixing or load-balancing.
\end{enumerate}

In summary, Attention, MoE, and DPR operate on orthogonal axes: Attention  governs \textit{which tokens to aggregate}, MoE \textit{which parameters to activate}, and DPR \textit{how to modulate each token's features}.
\section{Experiments}
\label{sec:experiments}

We evaluate DPR across twelve diverse benchmarks to answer five core research questions: 
\vspace{-0.5em}
\begin{itemize}[leftmargin=*]
\item \textbf{RQ1 (Performance)}: How does DPRNet compare with competitive baselines?
\vspace{-0.3em}
\item \textbf{RQ2 (Universality)}: Can DPR serve as a backbone-agnostic adapter across diverse backbones?
\vspace{-0.3em}
\item \textbf{RQ3 (Scaling)}: Does adaptive recalibration outperform naive scaling of static backbones?
\vspace{-0.3em}
\item \textbf{RQ4 (Ablations)}: How do components, hyperparameters, and efficiency impact DPR?
\vspace{-0.3em}
\item \textbf{RQ5 (Visualization)}: Can DPR visually reveal pattern-aware modulation and latent dynamics?
\end{itemize}
\vspace{-0.5em}

\begin{wrapfigure}{r}{0.54\linewidth}
  \centering
  \vspace{-1.5em}
  \includegraphics[width=\linewidth]{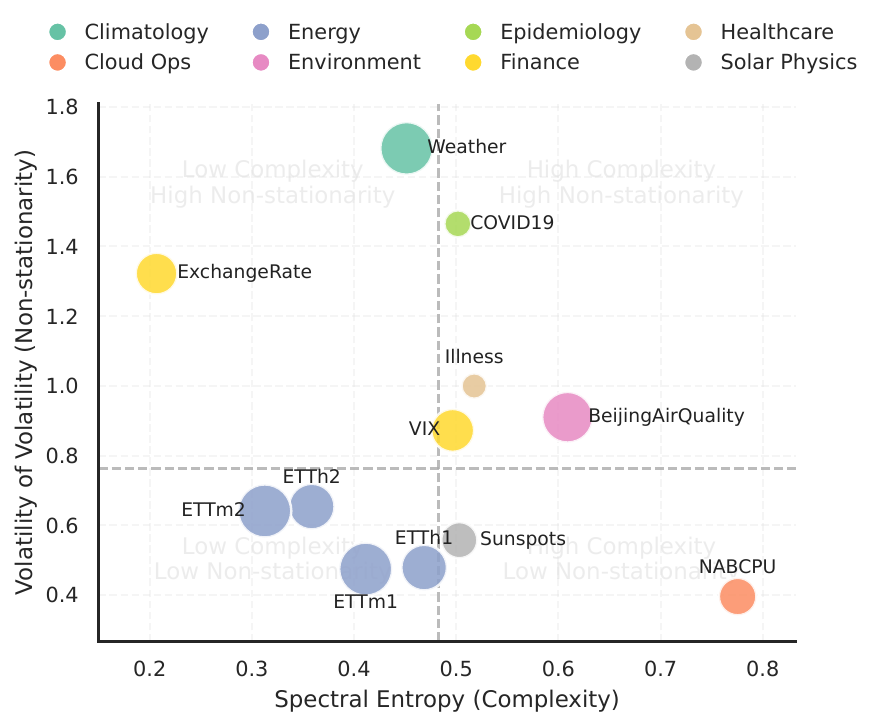}
  \caption{\textbf{Dataset Diversity Landscape.} Complexity vs. Non-stationarity; bubble size $\propto$ data volume.}
  \label{fig:dataset_landscape}
  \vspace{-1.5em}
\end{wrapfigure}

\textbf{Datasets and Non-stationarity Profiling.}  In contrast to the dominant practice of evaluating on datasets with pronounced seasonality and trend (e.g., ETT, Weather, Exchange), our benchmark suite incorporates high-complexity scenarios with irregular dynamics and weak periodic structure. We use twelve real-world benchmarks spanning eight domains (energy, finance, climatology, epidemiology, healthcare, cloud ops, environment, solar physics)~\citep{zhou2021informer,grigsby2021spacetimeformer}, covering diverse non-stationarity and dynamics (full provenance in Appendix~\ref{sec:appendix_data}). We profile each via Spectral Entropy ($H_s$, frequency complexity)~\citep{inouye1991quantification} and Volatility-of-Volatility (VoV, pattern shift intensity)~\citep{corsi2008volatility}. Figure~\ref{fig:dataset_landscape} reveals this diversity: Datasets distribute diversely across different complexity and non-stationarity regimes.

\textbf{Baselines and Evaluation Protocol.}  We compare eight models across diverse paradigms: attention-based architectures (Informer \citep{zhou2021informer}, Crossformer \citep{zhang2023crossformer}, iTransformer \citep{liu2024itransformer}, PatchTST \citep{nie2023patchtst}), efficient linear/MLP models enhanced by structural priors (TimeMixer \citep{wang2024timemixer}, WPMixer \citep{Murad_2025}), and complex filtering operations (TimesNet \citep{wu2023timesnet}, TimeFilter \citep{hu2025timefilter}). We exclude foundation models from the main tables as their massive proprietary pre-training renders from-scratch comparisons inequitable. We use MSE and MAE as evaluation metrics.

\textbf{Implementation Details.} All models follow a unified setup, with look-back $L$ and prediction horizon $H$ configured per dataset (full in Appendix~\ref{sec:appendix_impl}). We use the Adam optimizer \citep{kingma2014adam}. For adapter experiments (RQ2), we preserve baseline hyperparameters exactly and grid-search only DPR-specific parameters: basis size $K \in \{4, 8, 16\}$ and orthogonal penalty $\lambda_{\mathrm{orth}} \in \{0, 10^{-4}, 10^{-3}\}$. All experiments are conducted on 4 NVIDIA A800 GPUs.

\subsection{Main Results: Benchmark Performance (RQ1)}
\label{sec:main_results}

\textbf{Experimental Setup.} We benchmark DPRNet against eight state-of-the-art models across all 12 datasets. As described in Section~\ref{sec:method}, DPRNet uses a deliberately simplified linear backbone, ensuring that any improvement stems solely from the dynamic pattern recalibration of the DPR block. Full results across all prediction horizons are provided in Appendix~\ref{sec:appendix_results}.

\begin{table*}[h!]
    \scriptsize
    \centering
    \caption{\textbf{Main forecasting results of DPRNet on twelve benchmarks}. $MSE_{MAE}$ format with MAE in \textcolor[HTML]{666666}{gray}. \colorbox[RGB]{220, 237, 220}{\textbf{Bold}}/\colorbox[RGB]{255, 247, 205}{\underline{Underline}}: best/second best. Full results in Appendix~\ref{tab:main_results_full}.}
    \label{tab:main_results}
    
    \setlength{\tabcolsep}{1.5pt}
    \renewcommand{\arraystretch}{1.2} 
    
    \definecolor{bestbg}{RGB}{220, 237, 220}   
    \definecolor{secbg}{RGB}{255, 247, 205}    
    \definecolor{maegray}{HTML}{666666}        
    
    \newcommand{\bestbg}{\cellcolor{bestbg}}
    \newcommand{\secbg}{\cellcolor{secbg}}
    \newcommand{\ms}[2]{$#1_{\textcolor{maegray}{#2}}$}

    \resizebox{\textwidth}{!}{
    \begin{tabular}{@{}l c c c c c c c c c@{}}
    \toprule
    \multirow{2}{*}{\textbf{Dataset}} & 
    \makecell[c]{\textbf{Informer} \\ \citeyearpar{zhou2021informer}} & 
    \makecell[c]{\textbf{Crossformer} \\ \citeyearpar{zhang2023crossformer}} & 
    \makecell[c]{\textbf{TimesNet} \\ \citeyearpar{wu2023timesnet}} & 
    \makecell[c]{\textbf{PatchTST} \\ \citeyearpar{nie2023patchtst}} & 
    \makecell[c]{\textbf{iTransformer} \\ \citeyearpar{liu2024itransformer}} & 
    \makecell[c]{\textbf{TimeMixer} \\ \citeyearpar{wang2024timemixer}} &
    \makecell[c]{\textbf{TimeFilter} \\ \citeyearpar{hu2025timefilter}} &
    \makecell[c]{\textbf{WPMixer} \\ \citeyearpar{Murad_2025}} &
    \makecell[c]{\textbf{DPRNet} \\ \textbf{(Ours)}} \\
    \cmidrule(lr){2-2} \cmidrule(lr){3-3} \cmidrule(lr){4-4} \cmidrule(lr){5-5} \cmidrule(lr){6-6} \cmidrule(lr){7-7} \cmidrule(lr){8-8} \cmidrule(lr){9-9} \cmidrule(lr){10-10}
    
    \textbf{ILI} 
    & \ms{7.094}{1.896} & \ms{5.035}{1.542} & \ms{6.021}{1.309} & \ms{3.321}{1.110} & \ms{3.163}{1.069} & \ms{3.182}{1.148} 
    & \bestbg \ms{\mathbf{2.438}}{\mathbf{0.978}} & \ms{3.096}{1.069} 
    & \secbg \ms{\underline{2.963}}{\underline{1.088}} \\
    
    \textbf{BeijingAir} 
    & \ms{1.036}{0.740} & \bestbg \ms{\mathbf{0.424}}{\mathbf{0.422}} & \ms{0.537}{0.476} & \ms{0.451}{0.421} & \ms{0.457}{0.428} & \ms{0.448}{0.422} 
    & \secbg \ms{\underline{0.443}}{\underline{0.422}} & \ms{0.447}{0.424} 
    & \ms{0.451}{0.424} \\
    
    \textbf{COVID19} 
    & \ms{1.917}{0.632} & \ms{1.401}{0.476} & \ms{1.067}{0.538} & \secbg \ms{\underline{0.839}}{\underline{0.362}} & \ms{0.902}{0.379} & \ms{0.930}{0.401} 
    & \ms{0.908}{0.394} & \ms{0.839}{0.363} 
    & \bestbg \ms{\mathbf{0.804}}{\mathbf{0.366}} \\
    
    \textbf{Weather} 
    & \ms{0.484}{0.432} & \bestbg \ms{\mathbf{0.241}}{\mathbf{0.268}} & \ms{0.289}{0.305} & \ms{0.251}{0.270} & \ms{0.268}{0.280} & \ms{0.282}{0.301} 
    & \ms{0.246}{0.267} & \secbg \ms{\underline{0.243}}{\underline{0.264}} 
    & \ms{0.253}{0.272} \\
    
    \textbf{VIX} 
    & \ms{1.261}{0.744} & \ms{1.139}{0.685} & \ms{1.207}{0.715} & \ms{1.144}{0.692} & \ms{1.153}{0.699} & \ms{1.141}{0.694} 
    & \secbg \ms{\underline{1.115}}{\underline{0.677}} & \ms{1.119}{0.686} 
    & \bestbg \ms{\mathbf{1.108}}{\mathbf{0.682}} \\
    
    \textbf{NABCPU} 
    & \ms{2.689}{1.035} & \ms{1.257}{0.389} & \ms{1.220}{0.354} & \secbg \ms{\underline{1.195}}{\underline{0.318}} & \ms{1.200}{0.338} & \ms{1.196}{0.305} 
    & \ms{1.197}{0.307} & \ms{1.202}{0.334} 
    & \bestbg \ms{\mathbf{1.192}}{\mathbf{0.314}} \\
    
    \textbf{Sunspots} 
& \ms{2.562}{1.296} & \bestbg \ms{\mathbf{0.376}}{\mathbf{0.441}} & \ms{0.702}{0.578} & \ms{0.684}{0.576} & \ms{0.714}{0.587} & \ms{0.722}{0.586} 
& \ms{0.744}{0.590} & \secbg \ms{\underline{0.679}}{\underline{0.576}} 
& \ms{0.743}{0.603} \\
    
    \textbf{Exchange} 
    & \ms{2.782}{1.235} & \ms{0.894}{0.646} & \ms{0.494}{0.488} & \ms{0.453}{0.454} & \ms{0.446}{0.452} & \bestbg \ms{\mathbf{0.440}}{\mathbf{0.446}} 
    & \ms{0.446}{0.452} & \secbg \ms{\underline{0.445}}{\underline{0.449}} 
    & \bestbg \ms{\mathbf{0.440}}{\mathbf{0.446}} \\
    
    \textbf{ETTh1} 
    & \ms{1.324}{0.816} & \ms{0.449}{0.445} & \ms{0.534}{0.492} & \ms{0.459}{0.432} & \secbg \ms{\underline{0.448}}{\underline{0.431}} & \ms{0.458}{0.429} 
    & \ms{0.451}{0.429} & \ms{0.454}{0.431} 
    & \bestbg \ms{\mathbf{0.448}}{\mathbf{0.429}} \\
    
    \textbf{ETTh2} 
    & \ms{2.434}{1.074} & \ms{0.795}{0.599} & \ms{0.480}{0.460} & \ms{0.384}{0.403} & \ms{0.384}{0.400} & \ms{0.386}{0.402} 
    & \secbg \ms{\underline{0.380}}{\underline{0.402}} & \bestbg \ms{\mathbf{0.379}}{\mathbf{0.399}} 
    & \bestbg \ms{\mathbf{0.379}}{\mathbf{0.399}} \\
    
    \textbf{ETTm1} 
    & \ms{1.480}{0.855} & \ms{0.413}{0.404} & \ms{0.519}{0.474} & \ms{0.396}{0.387} & \ms{0.399}{0.390} & \secbg \ms{\underline{0.393}}{\underline{0.385}} 
    & \ms{0.395}{0.387} & \bestbg \ms{\mathbf{0.390}}{\mathbf{0.384}} 
    & \ms{0.396}{0.385} \\
    
    \textbf{ETTm2} 
& \ms{3.357}{1.128} & \ms{0.379}{0.395} & \ms{0.348}{0.364} & \ms{0.279}{0.319} & \ms{0.284}{0.321} & \secbg \ms{\underline{0.277}}{\underline{0.317}} 
& \ms{0.280}{0.321} & \bestbg \ms{\mathbf{0.274}}{\mathbf{0.316}} 
& \ms{0.282}{0.321} \\

    \bottomrule
    \end{tabular}}
    \vspace{-1em}
\end{table*}

\begin{table*}[t!]
    \small 
    \centering
    \caption{\textbf{Backbone-agnostic DPR enhancement}. \colorbox[RGB]{220, 237, 220}{Green} +DPR rows match or outperform Base. Win/Tie/Lose per backbone at bottom. Full results in Appendix~\ref{tab:plugin_results_full}.}
    \label{tab:plugin_results}
    
    \setlength{\tabcolsep}{5pt}  
    \renewcommand{\arraystretch}{1.1}
    
    \definecolor{bestbg}{RGB}{220, 237, 220}
    \definecolor{dprgray}{RGB}{247, 247, 247}
    \definecolor{maegray}{HTML}{666666}

    \newcommand{\cc}[2]{$#1_{\textcolor{maegray}{#2}}$} 
    \newcommand{\bb}[2]{$\mathbf{#1}_{\textcolor{maegray}{\mathbf{#2}}}$} 
    \newcommand{\ggb}[2]{\cellcolor{bestbg}$\mathbf{#1}_{\textcolor{maegray}{\mathbf{#2}}}$} 

    \resizebox{\textwidth}{!}{%
    \begin{tabular}{@{}ll c c c c c c c@{}}  
    \toprule
    \textbf{Dataset} & \textbf{Config} & \textbf{Informer} & \textbf{Crossformer} & \textbf{TimesNet} & \textbf{PatchTST} & \textbf{TimeMixer} & \textbf{TimeFilter} & \textbf{WPMixer} \\
    \midrule

    \textbf{ILI} 
    & Base & \cc{7.094}{1.896} & \cc{5.035}{1.542} & \cc{6.021}{1.309} & \cc{3.321}{1.110} & \cc{3.182}{1.148} & \cc{2.438}{0.978} & \cc{3.096}{1.069} \\ 
    \rowcolor{dprgray} \cellcolor{white}
    & +DPR & \ggb{5.740}{1.761} & \ggb{4.699}{1.466} & \ggb{4.335}{1.204} & \ggb{3.114}{1.095} & \ggb{3.172}{1.157} & \ggb{2.294}{0.951} & \ggb{2.965}{1.082} \\ 
    
    \textbf{BeijingAir} 
    & Base & \cc{1.036}{0.740} & \cc{0.424}{0.422} & \cc{0.537}{0.476} & \cc{0.451}{0.421} & \cc{0.448}{0.422} & \cc{0.443}{0.422} & \cc{0.447}{0.424} \\
    \rowcolor{dprgray} \cellcolor{white}
    & +DPR & \ggb{0.648}{0.574} & \ggb{0.417}{0.418} & \ggb{0.531}{0.473} & \ggb{0.449}{0.422} & \ggb{0.445}{0.421} & \ggb{0.442}{0.420} & \ggb{0.447}{0.424} \\
    
    \textbf{COVID19} 
    & Base & \cc{1.917}{0.632} & \cc{1.401}{0.476} & \cc{1.067}{0.538} & \cc{0.839}{0.362} & \cc{0.930}{0.401} & \cc{0.908}{0.394} & \cc{0.839}{0.363} \\
    \rowcolor{dprgray} \cellcolor{white}
    & +DPR & \ggb{1.742}{0.634} & \ggb{1.326}{0.476} & \ggb{1.035}{0.531} & \ggb{0.814}{0.359} & \ggb{0.864}{0.386} & \ggb{0.797}{0.374} & \ggb{0.786}{0.359} \\
    
    \textbf{Weather} 
    & Base & \cc{0.484}{0.432} & \cc{0.241}{0.268} & \cc{0.289}{0.305} & \cc{0.251}{0.270} & \cc{0.282}{0.301} & \cc{0.246}{0.267} & \cc{0.243}{0.264} \\
    \rowcolor{dprgray} \cellcolor{white}
    & +DPR & \ggb{0.386}{0.381} & \ggb{0.239}{0.267} & \ggb{0.285}{0.303} & \ggb{0.248}{0.269} & \ggb{0.243}{0.263} & \ggb{0.244}{0.265} & \ggb{0.243}{0.264} \\
    
    \textbf{VIX} 
    & Base & \cc{1.261}{0.744} & \cc{1.139}{0.685} & \cc{1.207}{0.715} & \bb{1.144}{0.692} & \cc{1.141}{0.694} & \cc{1.115}{0.677} & \cc{1.119}{0.686} \\
    \rowcolor{dprgray} \cellcolor{white}
    & +DPR & \ggb{0.939}{0.669} & \ggb{1.127}{0.686} & \ggb{1.134}{0.690} & \cc{1.146}{0.691} & \ggb{1.117}{0.683} & \ggb{1.094}{0.672} & \ggb{1.105}{0.679} \\
    
    \textbf{NABCPU} 
    & Base & \cc{2.689}{1.035} & \cc{1.257}{0.389} & \cc{1.220}{0.354} & \cc{1.195}{0.318} & \cc{1.196}{0.305} & \cc{1.197}{0.307} & \cc{1.202}{0.334} \\
    \rowcolor{dprgray} \cellcolor{white}
    & +DPR & \ggb{1.926}{0.777} & \ggb{1.236}{0.372} & \ggb{1.217}{0.352} & \ggb{1.194}{0.316} & \ggb{1.194}{0.300} & \ggb{1.195}{0.306} & \ggb{1.190}{0.312} \\
    
    \textbf{Sunspots} 
    & Base & \cc{2.562}{1.296} & \cc{0.376}{0.441} & \cc{0.702}{0.578} & \bb{0.684}{0.576} & \cc{0.722}{0.586} & \cc{0.744}{0.590} & \bb{0.679}{0.576} \\
    \rowcolor{dprgray} \cellcolor{white}
    & +DPR & \ggb{1.514}{1.001} & \ggb{0.357}{0.424} & \ggb{0.696}{0.575} & \cc{0.686}{0.578} & \ggb{0.689}{0.576} & \ggb{0.705}{0.578} & \ggb{0.672}{0.573} \\

    \textbf{Exchange} 
    & Base & \cc{2.782}{1.235} & \cc{0.894}{0.646} & \cc{0.494}{0.488} & \cc{0.453}{0.454} & \cc{0.440}{0.446} & \bb{0.446}{0.452} & \cc{0.445}{0.449} \\
    \rowcolor{dprgray} \cellcolor{white}
    & +DPR & \ggb{1.723}{0.948} & \ggb{0.840}{0.625} & \ggb{0.476}{0.480} & \ggb{0.448}{0.451} & \ggb{0.440}{0.446} & \cc{0.451}{0.452} & \ggb{0.442}{0.447} \\
    
    \textbf{ETTh1} 
    & Base & \cc{1.324}{0.816} & \cc{0.449}{0.445} & \cc{0.534}{0.492} & \bb{0.459}{0.432} & \cc{0.458}{0.429} & \bb{0.451}{0.429} & \cc{0.454}{0.431} \\
    \rowcolor{dprgray} \cellcolor{white}
    & +DPR & \ggb{1.000}{0.735} & \ggb{0.441}{0.440} & \ggb{0.531}{0.493} & \cc{0.460}{0.434} & \ggb{0.456}{0.428} & \cc{0.454}{0.430} & \ggb{0.453}{0.431} \\

    \textbf{ETTm1} 
    & Base & \cc{1.480}{0.855} & \cc{0.413}{0.404} & \bb{0.519}{0.474} & \cc{0.396}{0.387} & \cc{0.393}{0.385} & \cc{0.395}{0.387} & \cc{0.390}{0.384} \\
    \rowcolor{dprgray} \cellcolor{white}
    & +DPR & \ggb{1.097}{0.766} & \ggb{0.400}{0.397} & \cc{0.533}{0.480} & \ggb{0.395}{0.387} & \ggb{0.392}{0.385} & \ggb{0.394}{0.386} & \ggb{0.389}{0.384} \\
    \midrule

    \rowcolor{gray!15}
    \multicolumn{2}{l}{\textbf{DPR Win / Tie / Lose} } 
    & \textbf{10 / 0 / 0} 
    & \textbf{10 / 0 / 0} 
    & \textbf{9 / 0 / 1} 
    & \textbf{7 / 0 / 3} 
    & \textbf{9 / 1 / 0} 
    & \textbf{8 / 0 / 2} 
    & \textbf{8 / 2 / 0} \\

    \bottomrule
    \end{tabular}}
    \vspace{-1em}
\end{table*}

\textbf{Results Analysis.} Table~\ref{tab:main_results} shows DPRNet leading on volatile benchmarks (COVID19, VIX, NABCPU) where the static pattern response is most damaging. On more periodic datasets (ETT family, Exchange, Weather, Sunspots, BeijingAir) it trails only slightly on Weather, BeijingAir, Sunspots, and ETTm1/m2, while matching or winning horizon averages on Exchange and ETTh1/h2, all under a deliberately simple linear backbone that isolates DPR from architectural confounds. DPR targets the static pattern response directly, not a general capacity substitute. Consistent results across all twelve benchmarks validate this mechanism-level advantage.

\subsection{Backbone-Agnostic Enhancement (RQ2)}
\label{sec:universal_enhancement}

\textbf{Experimental Setup.} We further evaluate DPR as a backbone-agnostic adapter by integrating it into seven mainstream backbones (Informer, Crossformer, TimesNet, PatchTST, TimeMixer, TimeFilter, and WPMixer) and measuring the isolated gain from backbone-agnostic recalibration.

\textbf{Results Analysis.} DPR improves 61 of 70 backbone-dataset pairs (Table~\ref{tab:plugin_results}), with gains following two clear axes. Architecturally, attention-heavy models lacking token-level adaptation (Informer, Crossformer) benefit most; PatchTST, with inherent local patching, gains least. Per dataset, improvement concentrates on volatile benchmarks and attenuates on the stable ETT family, mirroring RQ1. The static pattern response is an uneven limitation: its severity depends on the backbone's local capacity and the data's non-stationarity. Even on the strongest baselines (iTransformer, TimeMixer), DPR yields consistent gains. It remedies the static pattern response systematically where it is needed most.

\subsection{Scaling Analysis: Parameter Scaling vs. DPR (RQ3)}
\label{sec:scaling_vs_dpr}

\textbf{Experimental Setup.} We assess whether scaling static backbones improves forecasting. Three backbones (PatchTST, TimesNet, TimeFilter) on ETTh1, ILI, and ExchangeRate are compared under six configurations: base, 2$\times$ depth, 2$\times$ width, 2$\times$ both, \textbf{+DPR}, and \textbf{Param Match (PM)}, the closest static variant at equal parameter count. FLOPs and parameters are profiled on ETTh1 via \texttt{thop}.

\begin{figure}[t]
    \centering
    \includegraphics[width=\textwidth]{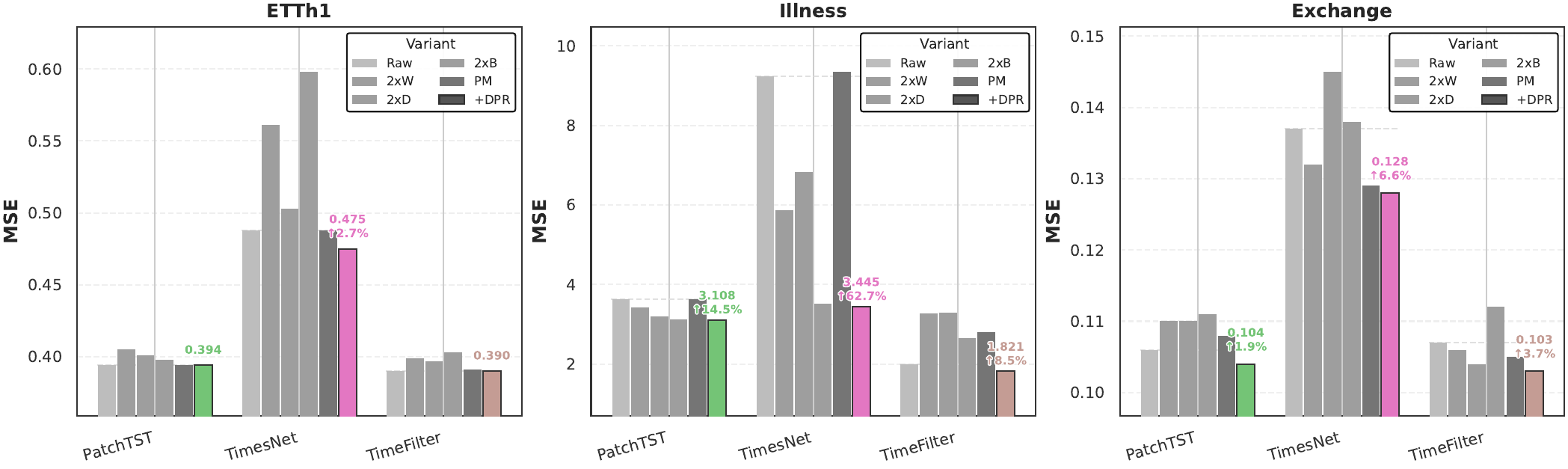}
    \caption{\textbf{Parameter Scaling vs.\ DPR.} Scaling backbone capacity often degrades performance. DPR achieves better gains at negligible cost. Full results in Appendix~\ref{tab:scaling_vs_dpr}.}
    \label{fig:scaling_vs_dpr}
\end{figure}
\begin{table*}[t!]
    \small
    \centering
    \caption{\textbf{Ablation of DPR Internal Mechanisms.} Full DPR shaded in gray. \textcolor[HTML]{2E7D32}{$\uparrow$}/\textcolor[HTML]{999999}{$\downarrow$}: MSE gain/drop vs.\ Baseline. \textbf{M-Scale}: multi-scale; \textbf{Ortho}: orthogonal reg.; \textbf{Init}: identity init.; \textbf{Route}: Soft vs.\ Hard.}

    \label{tab:ablation}

    \renewcommand{\arraystretch}{1}
    \setlength{\tabcolsep}{5pt} 

    \definecolor{bestbg}{RGB}{220, 237, 220}
    \definecolor{secbg}{RGB}{255, 247, 205}
    \definecolor{dprgray}{RGB}{245, 245, 245}
    \definecolor{maegray}{HTML}{666666}
    \definecolor{upgreen}{HTML}{2E7D32}
    \definecolor{dngray}{HTML}{999999}

    \newcommand{\bestbg}{\cellcolor{bestbg}}
    \newcommand{\secbg}{\cellcolor{secbg}}
    \newcommand{\dprbg}{\cellcolor{dprgray}}

    \newcommand{\ms}[2]{$#1_{\textcolor{maegray}{#2}}$}
    \newcommand{\bms}[2]{$\mathbf{#1}_{\textcolor{maegray}{\mathbf{#2}}}$}
    \newcommand{\sms}[2]{$\underline{#1}_{\textcolor{maegray}{\underline{#2}}}$}

    \newcommand{\up}[1]{\textcolor{upgreen}{$\uparrow$#1}}
    \newcommand{\dn}[1]{\textcolor{dngray}{$\downarrow$#1}}

    \resizebox{\textwidth}{!}{%
    \begin{tabular}{@{}l cccc ll ll ll@{}}
        \toprule
        \multirow{2}{*}{\textbf{Variants}} & \multicolumn{4}{c}{\textbf{Configurations}} & \multicolumn{2}{c}{\textbf{ETTh1}} & \multicolumn{2}{c}{\textbf{ILI}} & \multicolumn{2}{c}{\textbf{Exchange Rate}} \\
        \cmidrule(lr){2-5} \cmidrule(lr){6-7} \cmidrule(lr){8-9} \cmidrule(lr){10-11}
        & \textbf{M-Scale} & \textbf{Ortho} & \textbf{Init} & \textbf{Route} & \textbf{MSE}$_{\textcolor{maegray}{\textbf{MAE}}}$ & \textbf{Impv.} & \textbf{MSE}$_{\textcolor{maegray}{\textbf{MAE}}}$ & \textbf{Impv.} & \textbf{MSE}$_{\textcolor{maegray}{\textbf{MAE}}}$ & \textbf{Impv.} \\
        \midrule
        
        \textbf{Baseline} 
        & - & - & - & - 
        & \ms{0.394}{0.404} & - 
        & \ms{4.736}{1.480} & - 
        & \ms{0.269}{0.349} & - \\
        \midrule
        
        w/o M-Scale 
        & \ding{55} ($k=1$) & \ding{51} & \ding{51} & Soft 
        & \ms{0.397}{0.406} & \dn{0.8\%} 
        & \ms{5.129}{1.546} & \dn{8.3\%} 
        & \ms{0.266}{0.346} & \up{1.1\%} \\
        
        w/o Ortho
        & \ding{51} & \ding{55} & \ding{51} & Soft 
        & \ms{0.388}{0.401} & \up{1.5\%} 
        & \secbg \sms{4.661}{1.448} & \secbg \up{1.6\%} 
        & \ms{0.242}{0.338} & \up{10.0\%} \\
        
        w/o Init
        & \ding{51} & \ding{51} & \ding{55} (Rand) & Soft 
        & \ms{0.401}{0.410} & \dn{1.8\%} 
        & \ms{5.062}{1.525} & \dn{6.9\%} 
        & \ms{0.279}{0.367} & \dn{3.7\%} \\
        \midrule
        
        Discrete Routing
        & \ding{51} & \ding{51} & \ding{51} & \ding{55} \textbf{Hard} 
        & \secbg \sms{0.385}{0.399} & \secbg \up{2.3\%} 
        & \ms{4.721}{1.458} & \up{0.3\%} 
        & \secbg \sms{0.238}{0.335} & \secbg \up{11.5\%} \\
        \midrule
        
        \rowcolor{dprgray} \textbf{DPR (Ours)} 
        & \ding{51} & \ding{51} & \ding{51} & \ding{51} \textbf{Soft} 
        & \bestbg \bms{0.382}{0.397} & \bestbg \up{3.0\%} 
        & \bestbg \bms{4.593}{1.428} & \bestbg \up{3.0\%} 
        & \bestbg \bms{0.237}{0.335} & \bestbg \up{11.9\%} \\
        
        \bottomrule
    \end{tabular}%
    }
    \vspace{-1em}
\end{table*}
\begin{table}[t!]
    \small
    \centering
    \caption{\textbf{DPR vs.\ MoE Routing.} Identical DPRNet backbone. \textit{w/o DPR} removes the recalibration adapter; \textit{MoE} replaces soft-routing with discrete Top-$K$ gating ($K \in \{1,2,4\}$, 8 experts). Epoch training time and inference time in seconds, reported as train~s~/~inference~s.}
    \label{tab:moe_comparison}

    \renewcommand{\arraystretch}{1.15}
    \setlength{\tabcolsep}{3pt}

    \definecolor{bestbg}{RGB}{220, 237, 220}
    \definecolor{secbg}{RGB}{255, 247, 205}
    \definecolor{maegray}{HTML}{666666}
    \definecolor{dngray}{HTML}{999999}

    \providecommand{\bestbg}{\cellcolor{bestbg}}
    \providecommand{\secbg}{\cellcolor{secbg}}
    \providecommand{\ms}[2]{$#1_{\textcolor{maegray}{#2}}$}
    \providecommand{\bms}[2]{$\mathbf{#1}_{\textcolor{maegray}{\mathbf{#2}}}$}
    \providecommand{\sms}[2]{$\underline{#1}_{\textcolor{maegray}{\underline{#2}}}$}
    \providecommand{\dn}[1]{\textcolor{dngray}{$^{\downarrow\text{#1}}$}}

    \resizebox{\textwidth}{!}{%
    \begin{tabular}{@{}l ll ll ll ll ll ll@{}}
        \toprule
        \textbf{Model} & \multicolumn{2}{c}{\textbf{DPRNet}} & \multicolumn{2}{c}{\textbf{w/o DPR}} & \multicolumn{2}{c}{\textbf{MoE-Top1}} & \multicolumn{2}{c}{\textbf{MoE-Top2}} & \multicolumn{2}{c}{\textbf{MoE-Top4}} \\
        \cmidrule(lr){2-3} \cmidrule(lr){4-5} \cmidrule(lr){6-7} \cmidrule(lr){8-9} \cmidrule(lr){10-11}
        \textbf{Params} & \multicolumn{2}{c}{\scriptsize\textcolor{gray}{325K–602K}} & \multicolumn{2}{c}{\scriptsize\textcolor{gray}{287K–563K}} & \multicolumn{2}{c}{\scriptsize\textcolor{gray}{818K–1.1M}} & \multicolumn{2}{c}{\scriptsize\textcolor{gray}{818K–1.1M}} & \multicolumn{2}{c}{\scriptsize\textcolor{gray}{818K–1.1M}} \\
        \cmidrule(lr){2-3} \cmidrule(lr){4-5} \cmidrule(lr){6-7} \cmidrule(lr){8-9} \cmidrule(lr){10-11}
        \textbf{Dataset} & \textbf{MSE}$_{\textcolor{maegray}{\textbf{MAE}}}$ & \textbf{Time (s)} & \textbf{MSE}$_{\textcolor{maegray}{\textbf{MAE}}}$ & \textbf{Time (s)} & \textbf{MSE}$_{\textcolor{maegray}{\textbf{MAE}}}$ & \textbf{Time (s)} & \textbf{MSE}$_{\textcolor{maegray}{\textbf{MAE}}}$ & \textbf{Time (s)} & \textbf{MSE}$_{\textcolor{maegray}{\textbf{MAE}}}$ & \textbf{Time (s)} \\
        \midrule
        
        \textbf{Illness} (H=24)
        & \bestbg \bms{3.079}{1.096} & 0.85 / 0.73
        & \secbg \sms{3.347}{1.091} \dn{8.7\%} & 0.81 / 0.71
        & \ms{3.715}{1.156} \dn{20.7\%} & 1.06 / 0.77
        & \ms{3.755}{1.097} \dn{21.9\%} & 1.18 / 0.90
        & \ms{4.146}{1.122} \dn{34.7\%} & 1.02 / 0.77 \\
        
        \textbf{ETTh1} (H=96)
        & \bestbg \bms{0.392}{0.394} & 29.90 / 8.60
        & \secbg \sms{0.405}{0.398} \dn{3.4\%} & 15.84 / 5.07
        & \ms{0.411}{0.403} \dn{4.8\%} & 62.62 / 18.68
        & \ms{0.419}{0.405} \dn{6.8\%} & 89.12 / 16.05
        & \ms{0.417}{0.404} \dn{6.2\%} & 45.89 / 13.43 \\
        
        \bottomrule
    \end{tabular}%
    }
    \vspace{-1em}
\end{table}

\textbf{Results Analysis.} Figure~\ref{fig:scaling_vs_dpr} exposes the brittleness of naive scaling. The same static configuration that improves one dataset can degrade another, confirming that blind capacity inflation overfits dominant patterns rather than enabling adaptive response. DPR breaks this deadlock by design: it consistently outperforms or matches the PM static configuration across all backbones and datasets, and even at parity, does so at a fraction of the FLOPs. Adaptive recalibration and static parameter scaling are not substitutes—the former addresses a mechanism-level limitation the latter cannot cure by brute force.

\subsection{Ablations, Sensitivity, and Efficiency (RQ4)}
\label{sec:mechanisms_sensitivity}

\textbf{Ablation Studies.} Table~\ref{tab:ablation} evaluates four DPR variants isolating key design choices. The full DPR outperforms all ablated variants, forming a clear importance ranking. Identity initialization contributes most. Random initialization degrades performance globally, confirming the smooth optimization prior. Multi-scale perception delivers substantial gains on volatile datasets, where multi-resolution context matters most. Orthogonal regularization offers a smaller but consistent gain: basis diversification helps but is not the primary bottleneck. Discrete routing trails soft-routing on every dataset: discrete gating introduces routing discontinuity, validating soft-routing as a core design choice.

\textbf{MoE Architecture Comparison.} Replacing DPR's soft-routing with discrete MoE routing (Top-$K$, $K\!\in\!\{1,2,4\}$, 8 experts, load balancing) \citep{shazeer2017sparselygated,shi2025timemoe} adds more parameters (818K--1.1M), slows training, and degrades performance (Table~\ref{tab:moe_comparison}). Illness degrades sharply (Top-1: +20.7\% MSE, Top-4: +34.7\%); ETTh1 shows milder regression (+4.8\%--6.8\%). Simply removing DPR (w/o DPR) costs only 8.7\% and 3.4\%, confirming routing, not the adapter, is the bottleneck. Discrete routing incurs two penalties: load-balancing noise from expert selection, and restriction of each token to a fixed expert subset, blocking the continuous combination that soft-routing enables.

\begin{figure}[t!]
    \centering
    \vspace{-1em}
    \includegraphics[width=\linewidth]{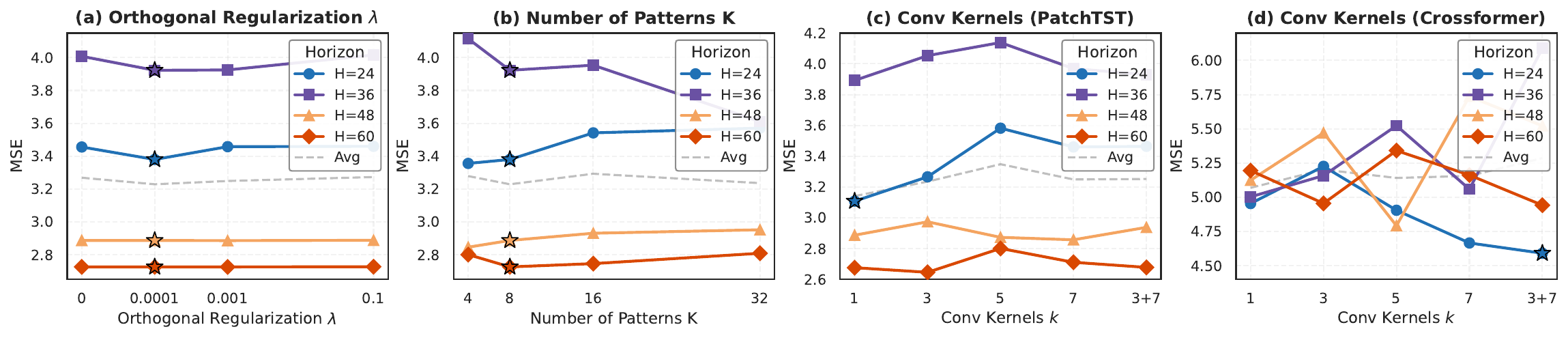}
    \caption{\textbf{Sensitivity Analysis.} (a) $\lambda_{\mathrm{orth}}$; (b) Basis size $K$; (c-d) Kernel configurations on PatchTST and Crossformer. Solid: per-horizon; dashed: horizon-averaged; $\star$: preferred setting.}
    \label{fig:sensitivity}
    \vspace{-2em}
\end{figure}

\textbf{Hyperparameter Sensitivity.} Evaluated on ILI (Fig.~\ref{fig:sensitivity}), we analyze basis size ($K$), orthogonal penalty ($\lambda_{\mathrm{orth}}$), and convolution kernels ($k$). Setting $\lambda_{\mathrm{orth}}=10^{-4}$ achieves the optimal trade-off. Increasing $K$ from 4 to 8 yields significant gains, confirming the need for basis diversity; however, performance plateaus for larger values, suggesting routing redundancy. Optimal kernels are architecture-dependent: PatchTST prefers pointwise filtering ($k=1$), whereas Crossformer favors multi-scale kernels ($k=(3,7)$), aligning with their respective inductive biases.

\begin{wrapfigure}{r}{0.45\linewidth}
  \centering
  \vspace{-1em}
  \includegraphics[width=\linewidth]{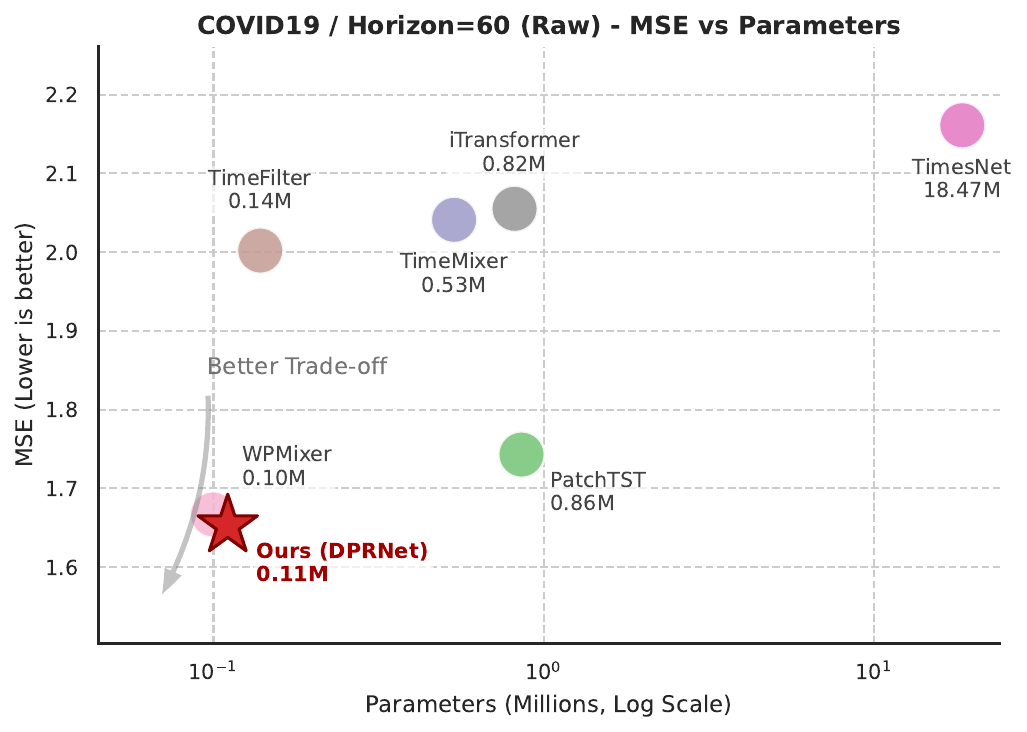}
  \caption{\textbf{Efficiency Trade-off.} Minimal error at negligible parameter cost.}
  \label{fig:efficiency_tradeoff}
  \vspace{-1em}
\end{wrapfigure}

\textbf{Computational and Parameter Efficiency.} Figure~\ref{fig:efficiency_tradeoff} maps the MSE-parameter trade-off on the highly non-stationary COVID19 dataset. Dense architectures suffer representation drift regardless of scale; their static mappings cannot escape pattern averaging. Minimalist DPRNet sits at the Pareto frontier, achieving the lowest error with a footprint comparable to the lightest baselines. Dynamic recalibration redefines the efficiency frontier. While blind scaling trades parameters for incremental gains along the same curve, DPR bends the curve, achieving lower error at negligible cost. Crucially, this requires no architectural bloat: the gain stems from the modulation mechanism, not from parameter count. Static budgets plateau; token-level recalibration transforms them.

\subsection{Visualization and Pattern Discovery (RQ5)}
\label{sec:qualitative_interpretability}

\begin{figure}[htbp]
  \centering
  \vspace{-0.5em}
  \includegraphics[width=\linewidth]{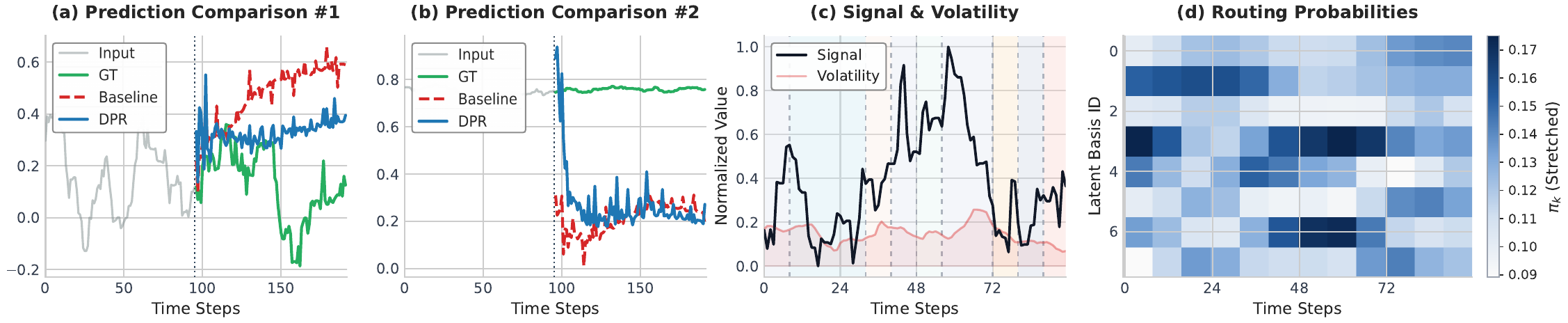}
  \caption{\textbf{Qualitative Evaluation.} \textbf{(a--b)} Under severe volatility, static backbones (red) drift while DPR (blue) closely tracks pattern shifts. \textbf{(c--d)} Routing probabilities concentrate stably during calm periods yet redistribute abruptly at volatility shocks (pink), uncovering latent response patterns.}
  \label{fig:interpretability}
  \vspace{-1em}
\end{figure}

Figure~\ref{fig:interpretability} contrasts DPR's adaptive tracking with static backbone drift. Panels (a--b) show static backbones (red) diverging from GT (green) in the forecast window, while DPR (blue) calibrates. Panel (c) traces routing-probability evolution, revealing pattern switching; panel (d) zooms in on volatility spikes. During calm periods, mass concentrates on smooth-trend bases; at spikes, it abruptly redistributes to transient-response bases, detecting regime boundaries without supervision.
 
\section{Conclusion and Future Work}
\label{sec:conclusion}

We identify and resolve the \textit{static pattern response} in deep time series forecasting, showing that effective forecasting demands shifting from naive parameter scaling to dynamic token-level recalibration: a mechanism-level solution, not an architectural patch. DPR validates this as a backbone-agnostic adapter that universally enhances static backbones by softly routing representations through learned adaptive response patterns and sits on the Pareto frontier of efficiency. 

\textbf{Limitations.} DPR relies exclusively on endogenous temporal features, leaving exogenous shocks undetected until they manifest in the signal. Future work will integrate multimodal covariates and embed DPR within large-scale pretrained models for anticipatory recalibration.

\bibliography{main}

\appendix

\clearpage
\startcontents[appendix]
\appendix

{\LARGE\bfseries Appendix}

\noindent\rule{\textwidth}{0.4pt}

{\large\bfseries Table of Contents}

\printcontents[appendix]{}{1}{}

\noindent\rule{\textwidth}{0.4pt}

\section{Dataset Statistics and Local Non-stationarity Analysis}
\label{sec:appendix_data}

We evaluate on \textbf{twelve} real-world benchmarks across eight diverse domains (energy, finance, climatology, healthcare, environment, epidemiology, cloud operations, and solar physics). These span from monthly (1-month) to sub-hourly (5-minute) resolutions, from 966 to 57{,}600 timesteps, and from univariate (1 variable) to moderately multivariate (21 variables), exhibiting a wide spectrum of non-stationarity, regime shifts, and volatility patterns. Summary statistics are in Table~\ref{tab:dataset_stats}.

\subsection{Dataset Descriptions}
\label{subsec:dataset_descriptions}

\begin{itemize}[leftmargin=*]
    \item \textbf{ETT (Electricity Transformer Temperature)}: Oil temperature and six loads from two Chinese counties (2016/7/1--2018/6/26). Four subsets: \textit{ETTh1/h2} (hourly, 14{,}400 steps) and \textit{ETTm1/m2} (15-min, 57{,}600 steps). Variables: HUFL, HULL, MUFL, MULL, LUFL, LULL, OT.

    \item \textbf{Weather}: 21 meteorological indicators (temperature, humidity, pressure, wind, precipitation, etc.) sampled every 10 minutes in 2020 (52{,}696 steps). Variables range from slowly drifting (temperature) to highly volatile (rain, wind gust).

    \item \textbf{Illness (ILI)}: Weekly ILI patient ratios from the U.S.~CDC (2002--2021, 966 steps). Seven variables: weighted/unweighted ILI ratios, age-specific counts, and total provider numbers. Small size and long-term dependencies with seasonal outbreaks.

    \item \textbf{ExchangeRate}: Daily exchange rates for eight countries (Australia, Britain, Canada, Switzerland, China, Japan, New Zealand, Singapore) from 1990 to 2016 (7{,}588 steps). Highly volatile, chaotic fluctuations with minimal periodicity challenge models that rely on regular cyclic patterns.

    \item \textbf{BeijingAirQuality}: Hourly air-quality indicators for Beijing (2014--2018, 36{,}000 steps). Variables: AQI, PM2.5, PM10, SO2, NO2, O3, CO (missing values linearly interpolated). Heterogeneous pollutants with non-stationary seasonal patterns and episodic haze events.

    \item \textbf{COVID19}: Confirmed COVID-19 cases from JHU CSSE (2020/1/22--2023/3/9, 1{,}143 steps). Following \citep{grigsby2021spacetimeformer}, daily new-cases for the top 7 countries plus Global aggregate (8 channels total). Asymmetric pattern shifts: long plateaus punctuated by exponential variant waves.

    \item \textbf{VIX (Volatility Index)}: Measures 30-day S\&P~500 implied volatility (1990/1/2--2026/4/15, 9{,}165 trading days, 1 channel). Canonical volatility clustering: narrow bands in calm markets, vertical spikes at crises (2008, 2020, 2022), then decay.

    \item \textbf{NABCPU}: AWS CloudWatch traces from the Numenta Anomaly Benchmark (2014/2/14--2014/2/28, 4{,}031 steps at 5-minute resolution, 3 channels). Sharp step-change bursts (scheduled jobs, memory leaks, traffic spikes) sit on top of irregular idle baselines, testing how well models recover from abrupt transitions.

    \item \textbf{Sunspots}: Monthly mean total sunspot number from WDC-SILSO, Royal Observatory of Belgium (1749/1--2026/3, 3{,}327 months, 1 channel). The $\sim$11-year cycle alternates between sharp spikes and smooth plateaus, producing very high VoV.
\end{itemize}

All datasets are split into train, validation, and test sets following standard protocols.

\begin{table}[h!]
    \centering
    \caption{Statistics of the twelve real-world benchmark datasets. For ETT, we adopt the truncated 20-month protocol introduced by Informer~\citep{zhou2021informer}; all other datasets use the full publicly available horizons. The datasets are grouped by their respective domains.}
    \label{tab:dataset_stats}
    
    \renewcommand{\arraystretch}{1.2}
    \setlength{\tabcolsep}{4pt}
    
    \resizebox{\linewidth}{!}{%
    \begin{tabular}{@{} l c c c c c c l @{}}
        \toprule
        \textbf{Dataset} & \textbf{Vars ($C$)} & \textbf{Timesteps ($T$)} & \textbf{Split} & \textbf{Frequency} & \textbf{Pred. Len. ($H$)} & \textbf{Period} & \textbf{Domain} \\
        \midrule
        
        ETTh1             & 7  & 14{,}400 & 6:2:2 & Hourly  & \{96, 192, 336, 720\} & 2016/7--2018/6  & \multirow{4}{*}{Energy} \\
        ETTh2             & 7  & 14{,}400 & 6:2:2 & Hourly  & \{96, 192, 336, 720\} & 2016/7--2018/6  & \\
        ETTm1             & 7  & 57{,}600 & 6:2:2 & 15 Min  & \{96, 192, 336, 720\} & 2016/7--2018/6  & \\
        ETTm2             & 7  & 57{,}600 & 6:2:2 & 15 Min  & \{96, 192, 336, 720\} & 2016/7--2018/6  & \\
        \midrule
        
        Weather           & 21 & 52{,}696 & 7:1:2 & 10 Min  & \{96, 192, 336, 720\} & 2020/1--2021/1  & Climatology \\
        BeijingAirQuality & 7  & 36{,}000 & 7:1:2 & Hourly  & \{96, 192, 336, 720\} & 2014--2018      & Environment \\
        \midrule
        
        Illness (ILI)     & 7  & 966      & 7:1:2 & Weekly  & \{24, 36, 48, 60\}    & 2002--2020      & Healthcare \\
        COVID19           & 8  & 1{,}143  & 7:1:2 & Daily   & \{7, 14, 28, 60\}     & 2020/1--2023/3  & Epidemiology \\
        \midrule
        
        ExchangeRate      & 8  & 7{,}588  & 7:1:2 & Daily   & \{96, 192, 336, 720\} & 1990--2016      & \multirow{2}{*}{Finance} \\
        VIX               & 1  & 9{,}165  & 7:1:2 & Daily   & \{96, 192, 336, 720\} & 1990--2026      & \\
        \midrule
        
        NABCPU            & 3  & 4{,}031  & 7:1:2 & 5 Min   & \{24, 48, 96, 192\}   & 2014/2          & Cloud Ops \\
        Sunspots          & 1  & 3{,}327  & 7:1:2 & Monthly & \{12, 24, 48, 96\}    & 1749--2026      & Solar Phys. \\
        
        \bottomrule
    \end{tabular}%
    }
\end{table}

\subsection{Quantifying Local Non-stationarity}
\label{sec:appendix_heterogeneity}

The qualitative characterization in Table~\ref{tab:dataset_stats} is complemented by three quantitative diagnostics that directly probe the \emph{local} pattern structure motivating our method. For each dataset we sample up to $16$ channels, $z$-normalize each channel, and report channel-average of the following measures:

\begin{itemize}[leftmargin=*]
    \item \textbf{ADF $p$-value.} Test regression $\Delta y_t = \alpha + \beta t + \gamma y_{t-1} + \sum_{i=1}^p \delta_i \Delta y_{t-i} + \varepsilon_t$ ($H_0: \gamma = 0$, unit root). Larger $p$ indicates failure to reject non-stationarity.
    \item \textbf{Spectral entropy $H_s \in [0,1]$.} $H_s = -\frac{1}{\log N_f}\sum_{f} P(f)\log P(f)$ where $P(f)$ is the normalized power spectrum (DC excluded). Values near $0$ signify concentrated, quasi-periodic dynamics; values near $1$ indicate broadband stochastic dynamics in which no single frequency dominates.
    \item \textbf{Volatility-of-Volatility (VoV).} $\text{VoV} = \frac{\operatorname{std}(\sigma_t)}{\operatorname{mean}(\sigma_t)}$ with $\sigma_t = \operatorname{std}(x_{t-w+1:t})$ the rolling window standard deviation ($w \approx 1$~day, capped at $256$ steps). Unlike ADF, which only probes first-order stationarity, VoV directly quantifies non-stationarity in the \emph{local dynamics themselves}: higher VoV means alternating calm and turbulent patterns, i.e., exactly the setting where a static weight-sharing backbone is forced into a compromised average representation.
\end{itemize}

\begin{table*}[t!]
    \small
    \centering
    \caption{\textbf{Local Non-stationarity Diagnostics.} Datasets are sorted by their overall non-stationarity \textbf{Score}. \textbf{Weather}, \textbf{COVID19}, and \textbf{Exchange} exhibit the highest VoV, while \textbf{Exchange}, \textbf{NABCPU}, and \textbf{Illness} show the most prominent unit-root tendencies (ADF $p > 0.05$). The \colorbox[RGB]{220, 237, 220}{\textbf{1st}} and \colorbox[RGB]{255, 247, 205}{\underline{2nd}} most non-stationary values per metric are highlighted to denote the most challenging datasets.}
    \label{tab:dataset_heterogeneity}
    
    \renewcommand{\arraystretch}{1.25}
    \setlength{\tabcolsep}{4.5pt}
    
    \definecolor{bestbg}{RGB}{220, 237, 220}
    \definecolor{secbg}{RGB}{255, 247, 205}
    
    \newcommand{\fst}[1]{\cellcolor{bestbg}\textbf{#1}}
    \newcommand{\snd}[1]{\cellcolor{secbg}\underline{#1}}
    
    \resizebox{\textwidth}{!}{%
    \begin{tabular}{@{} l cccccccccccc @{}}
        \toprule
        \textbf{Metric} & \textbf{Illness} & \textbf{BeijingAir} & \textbf{COVID19} & \textbf{Weather} & \textbf{VIX} & \textbf{NABCPU} & \textbf{Sunspots} & \textbf{Exchange} & \textbf{ETTh1} & \textbf{ETTh2} & \textbf{ETTm2} & \textbf{ETTm1} \\
        \midrule
        
        ADF $p$-val $\downarrow$ 
        & 0.0722 & 0.0000 & 0.0438 & 0.0033 & 0.0000 & \snd{0.1374} & 0.0002 & \fst{0.5499} & 0.0165 & 0.0249 & 0.0249 & 0.0165 \\
        $H_s$ 
        & 0.5176 & \snd{0.6089} & 0.5016 & 0.4514 & 0.4965 & \fst{0.7754} & 0.5030 & 0.2067 & 0.4686 & 0.3586 & 0.3126 & 0.4114 \\
        VoV $\uparrow$ 
        & 0.9995 & 0.9100 & \snd{1.4648} & \fst{1.6813} & 0.8722 & 0.3955 & 0.5573 & 1.3221 & 0.4786 & 0.6536 & 0.6411 & 0.4743 \\
        \midrule
        
        rank($H_s$) 
        & 10 & \snd{11} & 8  & 5  & 7 & \fst{12} & 9 & 1  & 6 & 3 & 2 & 4 \\
        rank(VoV)   
        & 9  & 8  & \snd{11} & \fst{12} & 7 & 1  & 4 & 10 & 3 & 6 & 5 & 2 \\
        \midrule
        
        Score 
        & \fst{19} & \fst{19} & \fst{19} & \snd{17} & 14 & 13 & 13 & 11 & 9 & 9 & 7 & 6 \\
        
        \bottomrule
    \end{tabular}%
    }
\end{table*}
The non-stationarity score in Table~\ref{tab:dataset_heterogeneity} is computed as $\textit{rank}(H_s) + \textit{rank}(\text{VoV})$, where higher values indicate more severe local non-stationarity that cannot be captured by a static global model.

\subsection{Observations and Discussion}
\label{subsec:observations}

The diagnostics confirm several patterns relevant to our approach.

\begin{enumerate}[leftmargin=*]
    \item \textbf{COVID19} and \textbf{Illness} (Score 19) are dominated by genuinely non-periodic dynamics. COVID19 has high VoV (1.46) from asymmetric pattern shifts: plateaus punctuated by exponential waves. Illness ranks high on both $H_s$ and VoV, reflecting episodic outbreaks with no fixed periodicity. Neither dataset offers a periodic backbone that a static model can rely on.
    \item \textbf{BeijingAirQuality} reaches the same composite Score (19) via a different mechanism. It has a strong daily and seasonal periodic base (rush-hour emission cycles, meteorological patterns), with episodic haze events layered on top as additive bursts. A periodic backbone captures most of the variance; the non-stationarity is real but secondary.
    \item \textbf{Weather} achieves VoV of 1.68, the highest overall, but each individual channel remains strongly periodic. The high VoV reflects cross-variable heterogeneity---smooth temperature alongside bursty precipitation and wind---rather than intrinsic unpredictability per channel.
    \item \textbf{VIX} (Score 14) shows volatility clustering: narrow low-volatility bands during calm periods with vertical spikes at crises (2008, 2020, 2022).
    \item \textbf{NABCPU} (Score 13) has the highest spectral entropy (0.78), indicating nearly broadband dynamics from overlapping periodicities: daily cycles, weekly patterns, and intermittent computational bursts. \textbf{Sunspots} (Score 13) shows the $\sim$11-year solar cycle with strong amplitude variation.
    \item \textbf{ExchangeRate} (Score 11) is the only dataset approaching a random walk (ADF $p=0.55$; the null of a unit root cannot be rejected at any conventional level). Its dynamics resemble a driftless stochastic process: the log-price today is approximately the log-price yesterday plus noise. This near-random-walk behaviour produces concentrated low-frequency power ($H_s=0.21$) yet, critically, high VoV (1.32), because the noise variance itself shifts over time (calm periods following major policy interventions). The result is a dataset that is statistically simple in expectation (persistent) but locally volatile in ways a fixed backbone cannot anticipate.
    \item \textbf{ETTh1/ETTh2/ETTm1/ETTm2} have the cleanest periodic structure (Score 6--9, low $H_s$, low VoV). These serve as a counterpoint where static backbones suffice and DPR should not degrade accuracy, which our main experiments confirm.
\end{enumerate}

\begin{figure}[h!]
    \centering
    \includegraphics[width=\linewidth]{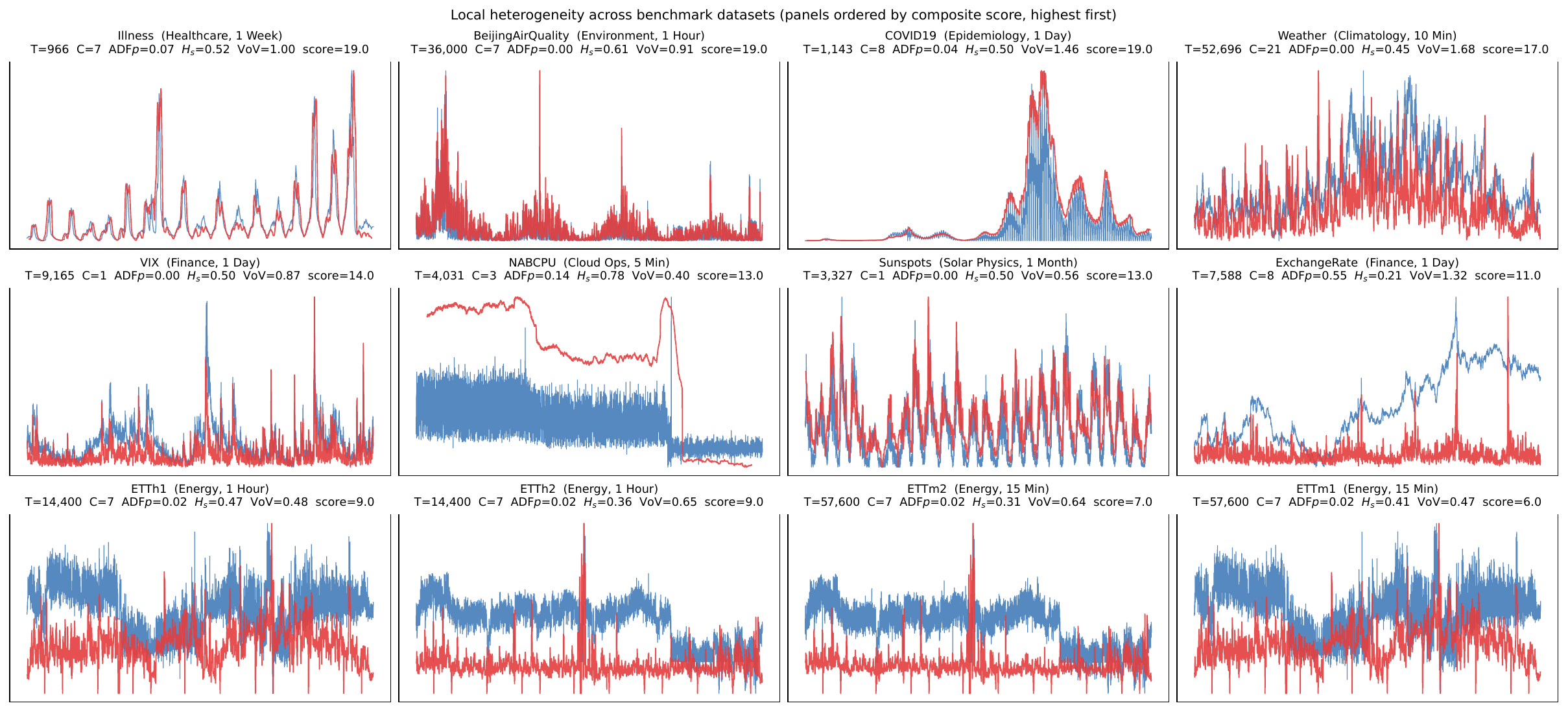}
    \caption{Local non-stationarity across the twelve benchmark datasets (4$\times$3 grid). For each dataset, a representative $z$-normalised channel (blue) is overlaid with its rolling standard deviation (red, right axis; window $\approx$ 1 day). A flat red curve indicates homogeneous local dynamics that a static weight-sharing backbone can absorb; a spiky red curve indicates pronounced pattern shifts that require adaptive pattern recalibration, which is precisely what DPR targets. Panel titles report length $T$, channels $C$, ADF $p$-value, spectral entropy $H_s$, and volatility-of-volatility VoV.}
    \label{fig:dataset_local_heterogeneity}
\end{figure}

Table~\ref{tab:dataset_heterogeneity} and Figure~\ref{fig:dataset_local_heterogeneity} show that local non-stationarity is not a corner case but a dominant property of real-world time series, motivating an adaptive recalibration mechanism like DPR.

\section{Implementation Details and Hyperparameters}
\label{sec:appendix_impl}

\subsection{DPR Default Configuration}
\label{subsec:dpr_config}

All models are implemented in PyTorch and trained on 4 NVIDIA A800 (80GB) GPUs. We use the Adam optimizer~\citep{kingma2014adam} across all experiments. When injecting DPR into existing backbones, their original hyperparameters (learning rate, dropout, batch size, layer count) are preserved exactly. Early stopping is applied with patience $10$ epochs based on validation loss. Table~\ref{tab:dpr_config} reports the complete default hyperparameters for our minimalist DPRNet backbone. Unless stated otherwise, adapter experiments adopt only the DPR-mechanism hyperparameters (bottom half of the table) while preserving the host backbone's native configuration.

\begin{table}[h!]
\centering
\caption{Default hyperparameters for DPRNet and the DPR mechanism.}
\label{tab:dpr_config}
\small
\setlength{\tabcolsep}{6pt}
\begin{tabularx}{\textwidth}{>{\raggedright\arraybackslash}X>{\centering\arraybackslash}X}
\toprule
\textbf{Hyperparameter} & \textbf{Default} \\
\midrule
\multicolumn{2}{l}{\textbf{Backbone}} \\
\quad Patch length $P$ / stride $S$ & $16$ / $8$ \\
\quad Hidden dimension $d$ & $256$ \\
\quad MLP: layers / expansion / dropout & $2$ / $2.0\times$ / $0.1$ \\
\midrule
\multicolumn{2}{l}{\textbf{DPR Mechanism}} \\
\quad Number of patterns $K$ & $8$ \\
\quad Projection $d_c$ & $\max(16, \lfloor d/4 \rfloor)$ \\
\quad Orthogonal penalty $\lambda_{\mathrm{orth}}$ & $10^{-4}$ \\
\quad Scale parameter $\tau$ & $1.0$ (learned) \\
\quad Gain init $\gamma$ & $0$ (identity) \\
\quad Multi-scale kernels & $k_1=3$, $k_2=7$ \\
\midrule
\multicolumn{2}{l}{\textbf{Training}} \\
\quad Optimizer / learning rate & Adam / $10^{-3}$ \\
\quad Batch size & $32$ \\
\quad Early-stopping patience & $10$ epochs \\
\bottomrule
\end{tabularx}
\end{table}

See Table~\ref{tab:dataset_stats} in Appendix~\ref{sec:appendix_data} for look-back windows and horizons.

\subsection{DPR Pseudocode}
\label{subsec:pseudocode}

Algorithm~\ref{alg:dpr} provides the complete forward pass.

\begin{algorithm}[h]
\caption{Dynamic Pattern Recalibration (DPR)}
\label{alg:dpr}
\begin{algorithmic}[1]
\Require Hidden states $\mathbf{H} \in \mathbb{R}^{B \times L \times d}$; basis $\mathbf{M} \in \mathbb{R}^{K \times d}$; centroids $\mathbf{E} \in \mathbb{R}^{K \times d_c}$; parameters $\mathbf{W}_1, \mathbf{b}_1, \tau, \gamma$; kernels $k_1, k_2$
\Ensure Modulated hidden states $\tilde{\mathbf{H}} \in \mathbb{R}^{B \times L \times d}$
\State $\mathbf{H}^{\top} \leftarrow \text{Permute}(\mathbf{H})$ \Comment{$B \times d \times L$}
\State $\mathbf{Z} \leftarrow \text{Concat}\bigl(\text{DWConv}_{k_1}(\mathbf{H}^{\top}), \text{DWConv}_{k_2}(\mathbf{H}^{\top})\bigr)^{\top}$ \Comment{$B \times L \times 2d$}
\State $\hat{\mathbf{E}}_k \leftarrow \mathbf{E}_k / \|\mathbf{E}_k\|_2, \ \forall k$ \Comment{Pre-compute normalized centroids}
\For{$b = 1, \dots, B$; $\ell = 1, \dots, L$}
    \State $\mathbf{c}_{b,\ell} \leftarrow \text{GELU}(\mathbf{W}_1 \mathbf{z}_{b,\ell} + \mathbf{b}_1)$ \Comment{$d_c$}
    \State $\hat{\mathbf{c}} \leftarrow \mathbf{c}_{b,\ell} / \|\mathbf{c}_{b,\ell}\|_2$ \Comment{$L_2$ norm}
    \State $\pi_{b,\ell,k} \leftarrow \text{softmax}\bigl(\tau \cdot \langle \hat{\mathbf{c}}, \hat{\mathbf{E}}_k \rangle\bigr)$ \Comment{Routing}
    \State $\mathbf{m}_{b,\ell} \leftarrow \sum_{k=1}^{K} \pi_{b,\ell,k} \, \mathbf{M}_{k,:}$ \Comment{Modulation vector}
    \State $\tilde{\mathbf{h}}_{b,\ell} \leftarrow \mathbf{h}_{b,\ell} \odot (\mathbf{1} + \gamma \mathbf{m}_{b,\ell})$ \Comment{Recalibration}
\EndFor
\State \Return $\tilde{\mathbf{H}}$
\end{algorithmic}
\end{algorithm}

\section{Integration Details for Diverse Architectures}
\label{sec:appendix_integration}

DPR is backbone-agnostic. We insert it at each backbone's late-stage hidden representation, before the final prediction head. For backbones without standard token-form hidden states, we apply lightweight reshaping before and after DPR. Table~\ref{tab:integration} summarizes the insertion points.
\begin{table}[h!]
\centering
\caption{DPR integration configuration across backbones (implementation-aligned).}
\label{tab:integration}
\small
\setlength{\tabcolsep}{4pt}
\begin{tabularx}{\textwidth}{l>{\raggedright\arraybackslash\hsize=0.7\hsize}X>{\raggedright\arraybackslash\hsize=1.3\hsize}Xcc}
\toprule
\textbf{Backbone} & \textbf{Hidden Shape} & \textbf{Insert Position} & \textbf{Reshape?} & \textbf{Kernel} \\
\midrule
Informer & $B \times L \times d$ & After decoder, before head & No & $(3,7)$ \\
Crossformer & $(B\!\cdot\!N)\times P_{\text{out}}\times d$ & Decoder: final scale branch & No & $(3,7)$ \\
TimesNet & $B \times (L{+}H) \times d$ & After backbone, before head & No & $(3,7)$ \\
PatchTST & $B \times N \times P \times d$ & After encoder, before head & Yes & $1$ \\
TimeMixer & $B \times H \times d$ (per scale) & Per-scale: after temporal proj. & No & $(3,7)$ \\
TimeFilter & $B \times N \times P \times d$ & After graph backbone, before head & Yes & $1$ \\
WPMixer & $B \times C \times P \times d$ & Per-resolution: after mixer, before head & Yes & $1$ \\
\bottomrule
\end{tabularx}
\end{table}

For sequential token topologies (Informer, TimesNet, and the per-scale path of TimeMixer), DPR consumes hidden states in $\mathbb{R}^{B\times L\times d}$ (or forecast-length variants) and applies depthwise convolutions without reshaping. A single DPR adapter is shared per backbone, applied at late-stage representations (single point or per-scale/per-branch reuse). We do not stack independent DPR adapters.

\subsection{Patch/Channel-Aware Topologies}
\label{subsec:patch_topologies}
For PatchTST and TimeFilter, hidden states are organized as $\mathbb{R}^{B\times N\times P\times d}$. We reshape to
\begin{equation}
\mathbf{H}_{\text{in}} \in \mathbb{R}^{B \times N \times P \times d}
\;\xrightarrow{\text{reshape}}\;
\hat{\mathbf{H}}_{\text{in}} \in \mathbb{R}^{(B\cdot N)\times P\times d},
\end{equation}
apply DPR independently on each variable-token sequence, then restore the original layout.

\subsection{WPMixer Branch-Level Integration}
\label{subsec:wpmixer_integration}
In WPMixer, DPR is applied inside each resolution branch. The branch feature map is reshaped from $\mathbb{R}^{B\times C\times P\times d}$ to $\mathbb{R}^{B\times(CP)\times d}$, modulated, and reshaped back before the branch prediction head.

\subsection{Kernel Configuration}
\label{subsec:kernel_config}
The DPR adapter defaults to multi-scale kernels $(3,7)$. In our baseline experiment scripts, we set $k=1$ for patch-like backbones (PatchTST, TimeFilter, WPMixer) as a task-specific choice, while keeping $(3,7)$ for the others.

\section{Inductive Bias Analysis of DPR}
\label{sec:appendix_inductive_bias}

We analyze why DPR works for adaptive time series forecasting and why static capacity scaling cannot replicate token-level modulation.

\subsection{From Static Capacity to Dynamic Gain Control}
\label{subsec:static_dynamic}

Consider a linear transformation $\mathcal{F}_{\Theta}(\mathbf{h}) = \mathbf{W}\mathbf{h}$ with $\mathbf{W} \in \mathbb{R}^{d \times d}$. Its Jacobian is $\mathbf{W}$, a constant applied identically to every token. Widening $\mathbf{W}$ or stacking layers does not change this: the Jacobian stays fixed across temporal positions. The model learns a compromised average representation, unable to adopt conservative gains for stable patterns and high for volatile shocks.
    
DPR resolves this via $\tilde{\mathbf{h}} = \mathbf{h} \odot (\mathbf{1} + \gamma\,\mathbf{m})$, where $\mathbf{m}$ is a context-weighted combination of learned basis vectors. Dropping the $\mathbf{m}(\mathbf{h})$ dependence, the Jacobian becomes approximately diagonal:
\begin{equation}
\frac{\partial \tilde{\mathbf{h}}}{\partial \mathbf{h}} \approx \operatorname{diag}(\mathbf{1} + \gamma\,\mathbf{m}).
\label{eq:jacobian_diag_appendix}
\end{equation}
Each token receives personalized, temporally conditioned scaling of feature dimensions. Under stable patterns, $\mathbf{m}$ stays near zero, keeping the Jacobian close to identity. When a volatile shock is detected, certain dimensions of $\mathbf{m}$ activate, increasing gain on those feature axes. No static parameter matrix, regardless of size, can replicate this token-level, context-conditioned stretching.

\subsection{Design Principles: Continuity, Orthogonality, and Information Capacity}
\label{subsec:design_principles}

\textbf{Continuity bias via soft routing.} Pattern transitions in physical systems are typically gradual. With soft routing (softmax over pattern logits), the convex combination $\mathbf{m}_{b,\ell} = \sum_k \pi_{b,\ell,k}\,\mathbf{M}_{k,:}$ varies smoothly with the local context $\mathbf{z}_{b,\ell}$, so small context changes induce small modulation changes. This aligns with the temporal coherence of real-world dynamics.

\textbf{Orthogonal basis as a compact coordinate system.} The regularizer $\mathcal{L}_{\mathrm{orth}}$ penalizes Gram-matrix deviation from identity, encouraging near-orthogonal pattern vectors. This discourages redundant modes and promotes disentangled response directions: separate basis vectors for ``steady trend'' versus ``spike-and-decay'' patterns, rather than collapsing to a single averaged template. A compact, diverse basis also improves interpretability.

\textbf{Information-theoretic view.} Suppose the routing distribution $\boldsymbol{\pi}$ over $K$ basis vectors is obtained by softmax projection of inner products with centroids $\mathbf{E}$. If the basis vectors $\mathbf{M}_{k,:}$ are highly correlated, distinct centroids map to nearly identical modulation directions, making routing decisions uninformative. The Gram matrix $\mathbf{G}$ then has off-diagonal entries close to $1$. Minimizing $\|\mathbf{G} - \mathbf{I}\|_F^2$ pushes $\mathbf{G}$ toward identity, maximizing $\det(\mathbf{G})$, which measures the information capacity of the basis. A larger determinant means more independent modulation directions and more informative routing decisions. Orthogonal regularization ensures each additional basis vector contributes a genuinely independent modulation direction.

\subsection{Formal Theoretical Guarantees: Full Jacobian and Lipschitz Stability}
\label{subsec:theory_guarantees}

Equation~\eqref{eq:jacobian_diag_appendix} ignores the dependence of $\mathbf{m}$ on $\mathbf{h}$ through the perception network. Consider the full Jacobian. Let $\mathbf{s} = \mathbf{1} + \gamma\,\mathbf{m}$ denote the scaling vector. The $(i,j)$-th element of the total derivative is
\begin{equation}
\frac{\partial \tilde{h}_i}{\partial h_j} = s_i \, \delta_{ij} + h_i \, \frac{\partial s_i}{\partial h_j},
\label{eq:full_jacobian}
\end{equation}
where $\delta_{ij}$ is the Kronecker delta. In matrix form, $\partial \tilde{\mathbf{h}}/\partial \mathbf{h} = \operatorname{diag}(\mathbf{s}) + \operatorname{diag}(\mathbf{h}) \, (\partial \mathbf{s}/\partial \mathbf{h})$. Since $\mathbf{s} = \mathbf{1} + \gamma\,\mathbf{m}$, we have $\partial \mathbf{s}/\partial \mathbf{h} = \gamma \, (\partial \mathbf{m}/\partial \mathbf{h})$, so the off-diagonal term is proportional to $\gamma$. With $\gamma$ initialized at $0$ and growing, this second term is zero at initialization and stays small relative to the diagonal term. Thus $\operatorname{diag}(\mathbf{s})$ dominates, and the token-level gain interpretation in Eq.~\eqref{eq:jacobian_diag_appendix} is accurate.

\textbf{Lipschitz stability.} A concern with adaptive mechanisms is whether they destabilize the forward pass or amplify noise. Since $\tilde{\mathbf{h}} = \mathbf{h} \odot (\mathbf{1} + \gamma\,\mathbf{m})$ and each component of $\mathbf{m}$ is a convex combination of basis vectors with coefficients in $[0,1]$, we have
\begin{equation}
\|\tilde{\mathbf{h}}\|_2 \;\le\; \max_i |1 + \gamma\,m_i| \cdot \|\mathbf{h}\|_2 \;\le\; (1 + \gamma\,\max_k \|\mathbf{M}_{k,:}\|_\infty) \cdot \|\mathbf{h}\|_2.
\end{equation}
With $\gamma$ initialized at $0$ and learned, the effective Lipschitz constant of DPR stays close to $1$ at the start and remains bounded throughout training. This contrasts with unconstrained affine adapters (e.g., standard FiLM without residual gating), which can amplify or suppress features and suffer training instability in deep stacks. The residual form $\mathbf{1} + \gamma\,\mathbf{m}$ serves dual purpose: it preserves backbone dynamics early in training and guarantees bounded amplification after recalibration activates.
\section{Complete Experimental Results}
\label{sec:appendix_results}

This section reports full per-horizon results, adapter integration gains, and scaling comparisons that complement the averaged tables in the main paper.

\subsection{Full Main Results}
Table~\ref{tab:main_results_full} reports MSE and MAE for all baselines across every prediction horizon. The per-horizon tables reinforce patterns that horizon averages compress but do not invent. Datasets where DPRNet ranks at the top on the average (COVID19, VIX, NABCPU) tend to show the same advantage at \emph{most} individual forecast lengths, so gains are not an artifact of aggregation alone. On COVID19 the recalibration margin often widens with the forecast horizon: DPRNet is competitive at 7-day horizons but leads more clearly at longer horizons (e.g., 60 days), consistent with static mappings drifting as look-ahead increases.

\begin{table*}[t!]
    \small
    \centering
    \caption{Full forecasting results on 12 public benchmarks. Lower values indicate better performance. We adopt the $MSE_{MAE}$ format to report both metrics compactly, where MAE is distinguished in \textcolor[HTML]{666666}{gray}. \colorbox[RGB]{220, 237, 220}{\textbf{Bold}} indicates the best result, and \colorbox[RGB]{255, 247, 205}{\underline{Underline}} indicates the second best.}
    \label{tab:main_results_full}

    \setlength{\tabcolsep}{3.5pt}
    \renewcommand{\arraystretch}{0.2}

    \definecolor{bestbg}{RGB}{220, 237, 220}
    \definecolor{secbg}{RGB}{255, 247, 205}
    \definecolor{maegray}{HTML}{666666}

    \newcommand{\bestbg}{\cellcolor{bestbg}}
    \newcommand{\secbg}{\cellcolor{secbg}}
    \newcommand{\ms}[2]{$#1_{\textcolor{maegray}{#2}}$}

    \resizebox{0.95\textwidth}{!}{%
    \begin{tabular}{@{}l l c c c c c c c c c @{}}
    \toprule
    \multirow{2}{*}{\textbf{Dataset}} & \multirow{2}{*}{\textbf{Len.}} & \multicolumn{1}{c}{\textbf{Informer}} & \multicolumn{1}{c}{\textbf{Crossformer}} & \multicolumn{1}{c}{\textbf{TimesNet}} & \multicolumn{1}{c}{\textbf{PatchTST}} & \multicolumn{1}{c}{\textbf{iTransformer}} & \multicolumn{1}{c}{\textbf{TimeMixer}} & \multicolumn{1}{c}{\textbf{TimeFilter}} & \multicolumn{1}{c}{\textbf{WPMixer}} & \multicolumn{1}{c}{\textbf{DPRNet}} \\
    \cmidrule(lr){3-3} \cmidrule(lr){4-4} \cmidrule(lr){5-5} \cmidrule(lr){6-6} \cmidrule(lr){7-7} \cmidrule(lr){8-8} \cmidrule(lr){9-9} \cmidrule(lr){10-10} \cmidrule(lr){11-11}
    &  & Base & Base & Base & Base & Base & Base & Base & Base & \textbf{(Ours)} \\
    \midrule
    \multirow{5}*{\rotatebox{90}{ILI}} & 24 & \ms{7.005}{1.868} & \ms{4.736}{1.480} & \ms{9.241}{1.389} & \ms{3.633}{1.079} & \ms{3.507}{1.071} & \ms{3.124}{1.136} & \bestbg \ms{\mathbf{1.991}}{\mathbf{0.873}} & \ms{3.173}{1.022}     & \secbg \ms{\underline{2.970}}{\underline{1.071}} \\
      & 36 & \ms{7.201}{1.898} & \ms{5.153}{1.561} & \ms{7.371}{1.438} & \ms{4.019}{1.192} & \ms{3.974}{1.152} & \ms{3.538}{1.214} & \bestbg \ms{\mathbf{2.481}}{\mathbf{0.976}} & \ms{3.720}{1.147}     & \secbg \ms{\underline{3.349}}{\underline{1.149}} \\
      & 48 & \ms{7.030}{1.903} & \ms{5.244}{1.576} & \ms{4.175}{1.237} & \ms{2.939}{1.099} & \secbg \ms{\underline{2.513}}{\underline{1.005}} & \ms{3.055}{1.130} & \bestbg \ms{\mathbf{2.468}}{\mathbf{0.990}} & \ms{2.770}{1.061}     & \ms{2.841}{1.081} \\
      & 60 & \ms{7.140}{1.916} & \ms{5.006}{1.550} & \ms{3.298}{1.172} & \ms{2.695}{1.071} & \bestbg \ms{\mathbf{2.657}}{\mathbf{1.049}} & \ms{3.010}{1.114} & \ms{2.812}{1.072} & \ms{2.722}{1.044} & \secbg \ms{\underline{2.692}}{\underline{1.052}} \\
    \cmidrule{2-11}
      & \textbf{Avg} & \ms{7.094}{1.896} & \ms{5.035}{1.542} & \ms{6.021}{1.309} & \ms{3.321}{1.110} & \ms{3.163}{1.069} & \ms{3.182}{1.148} & \bestbg \ms{\mathbf{2.438}}{\mathbf{0.978}} & \ms{3.096}{1.069} & \secbg \ms{\underline{2.963}}{\underline{1.088}} \\
    \midrule
    \multirow{5}*{\rotatebox{90}{BeijingAir}} & 96 & \ms{1.550}{1.047} & \ms{0.407}{0.418} & \ms{0.573}{0.491} & \secbg \ms{\underline{0.404}}{\underline{0.397}} & \ms{0.411}{0.406} & \ms{0.407}{0.401} & \ms{0.404}{0.404} & \bestbg \ms{\mathbf{0.403}}{\mathbf{0.402}} & \ms{0.408}{0.402} \\
      & 192 & \ms{1.081}{0.759} & \bestbg \ms{\mathbf{0.412}}{\mathbf{0.413}} & \ms{0.553}{0.485} & \ms{0.439}{0.415} & \ms{0.447}{0.423} & \ms{0.438}{0.417} & \secbg \ms{\underline{0.431}}{\underline{0.416}} & \ms{0.438}{0.419} & \ms{0.441}{0.420} \\
      & 336 & \ms{0.756}{0.579} & \bestbg \ms{\mathbf{0.433}}{\mathbf{0.424}} & \ms{0.502}{0.461} & \ms{0.466}{0.430} & \ms{0.471}{0.435} & \ms{0.461}{0.429} & \secbg \ms{\underline{0.455}}{\underline{0.426}} & \ms{0.458}{0.431} & \ms{0.464}{0.432} \\
      & 720 & \ms{0.757}{0.577} & \bestbg \ms{\mathbf{0.443}}{\mathbf{0.432}} & \ms{0.519}{0.467} & \ms{0.494}{0.443} & \ms{0.498}{0.449} & \ms{0.487}{0.441} & \secbg \ms{\underline{0.482}}{\underline{0.441}} & \ms{0.490}{0.446} & \ms{0.492}{0.444} \\
    \cmidrule{2-11}
      & \textbf{Avg} & \ms{1.036}{0.740} & \bestbg \ms{\mathbf{0.424}}{\mathbf{0.422}} & \ms{0.537}{0.476} & \ms{0.451}{0.421} & \ms{0.457}{0.428} & \ms{0.448}{0.422} & \secbg \ms{\underline{0.443}}{\underline{0.422}} & \ms{0.447}{0.424} & \ms{0.451}{0.424} \\
    \midrule
    \multirow{5}*{\rotatebox{90}{COVID19}}       & 7 & \ms{0.824}{0.391} & \ms{0.609}{0.295} & \ms{0.490}{0.390} & \ms{0.335}{0.216} & \bestbg \ms{\mathbf{0.327}}{\mathbf{0.217}} & \ms{0.361}{0.238} & \secbg \ms{\underline{0.331}}{\underline{0.220}} & \ms{0.343}{0.218} & \ms{0.327}{0.224} \\
      & 14 & \ms{0.849}{0.433} & \ms{0.853}{0.356} & \ms{0.690}{0.484} & \ms{0.440}{0.264} & \secbg \ms{\underline{0.439}}{\underline{0.269}} & \ms{0.464}{0.288} & \bestbg \ms{\mathbf{0.434}}{\mathbf{0.275}} & \ms{0.473}{0.267} & \ms{0.440}{0.275} \\
      & 28 & \ms{3.027}{0.813} & \ms{1.444}{0.531} & \ms{0.927}{0.550} & \ms{0.837}{0.387} & \bestbg \ms{\mathbf{0.787}}{\mathbf{0.385}} & \ms{0.855}{0.410} & \ms{0.866}{0.439} & \ms{0.872}{0.397} & \secbg \ms{\underline{0.813}}{\underline{0.396}} \\
      & 60 & \ms{2.970}{0.893} & \ms{2.700}{0.724} & \ms{2.161}{0.728} & \ms{1.743}{0.582} & \ms{2.055}{0.643} & \ms{2.041}{0.668} & \ms{2.002}{0.643} & \secbg \ms{\underline{1.667}}{\underline{0.569}} & \bestbg \ms{\mathbf{1.637}}{\mathbf{0.569}} \\
    \cmidrule{2-11}
      & \textbf{Avg} & \ms{1.917}{0.632} & \ms{1.401}{0.476} & \ms{1.067}{0.538} & \secbg \ms{\underline{0.839}}{\underline{0.362}} & \ms{0.902}{0.379} & \ms{0.930}{0.401} & \ms{0.908}{0.394} & \ms{0.839}{0.363} & \bestbg \ms{\mathbf{0.804}}{\mathbf{0.366}} \\
    \midrule
    \multirow{5}*{\rotatebox{90}{Weather}}       & 96 & \ms{1.094}{0.753} & \ms{0.166}{0.204} & \ms{0.181}{0.227} & \ms{0.173}{0.208} & \ms{0.185}{0.217} & \ms{0.180}{0.226} & \bestbg \ms{\mathbf{0.161}}{\mathbf{0.200}} & \secbg \ms{\underline{0.163}}{\underline{0.201}} & \ms{0.172}{0.209} \\
      & 192 & \ms{0.294}{0.328} & \bestbg \ms{\mathbf{0.208}}{\mathbf{0.244}} & \ms{0.246}{0.283} & \ms{0.215}{0.248} & \ms{0.235}{0.258} & \ms{0.246}{0.282} & \ms{0.211}{0.245} & \secbg \ms{\underline{0.208}}{\underline{0.242}} & \ms{0.217}{0.249} \\
      & 336 & \bestbg \ms{\mathbf{0.260}}{\mathbf{0.312}} & \secbg \ms{\underline{0.260}}{\underline{0.286}} & \ms{0.311}{0.324} & \ms{0.269}{0.288} & \ms{0.291}{0.299} & \ms{0.309}{0.323} & \ms{0.268}{0.286} & \ms{0.262}{0.281} & \ms{0.272}{0.290} \\
      & 720 & \bestbg \ms{\mathbf{0.289}}{\mathbf{0.336}} & \secbg \ms{\underline{0.331}}{\underline{0.339}} & \ms{0.417}{0.387} & \ms{0.345}{0.337} & \ms{0.362}{0.345} & \ms{0.395}{0.375} & \ms{0.346}{0.337} & \ms{0.340}{0.332} & \ms{0.351}{0.341} \\
    \cmidrule{2-11}
      & \textbf{Avg} & \ms{0.484}{0.432} & \bestbg \ms{\mathbf{0.241}}{\mathbf{0.268}} & \ms{0.289}{0.305} & \ms{0.251}{0.270} & \ms{0.268}{0.280} & \ms{0.282}{0.301} & \ms{0.246}{0.267} & \secbg \ms{\underline{0.243}}{\underline{0.264}} & \ms{0.253}{0.272} \\
    \midrule
    \multirow{5}*{\rotatebox{90}{VIX}}       & 96 & \ms{1.077}{0.668} & \ms{1.023}{0.571} & \ms{1.005}{0.562} & \secbg \ms{\underline{0.942}}{\underline{0.539}} & \ms{0.990}{0.560} & \ms{0.967}{0.550} & \ms{0.959}{0.545} & \ms{0.957}{0.547} & \bestbg \ms{\mathbf{0.935}}{\mathbf{0.542}} \\
      & 192 & \ms{2.059}{0.975} & \ms{1.245}{0.683} & \ms{1.275}{0.693} & \ms{1.268}{0.692} & \ms{1.208}{0.674} & \ms{1.203}{0.670} & \bestbg \ms{\mathbf{1.181}}{\mathbf{0.662}} & \secbg \ms{\underline{1.193}}{\underline{0.665}} & \ms{1.193}{0.668} \\
      & 336 & \bestbg \ms{\mathbf{1.011}}{\mathbf{0.700}} & \ms{1.240}{0.769} & \ms{1.342}{0.802} & \ms{1.224}{0.762} & \ms{1.176}{0.751} & \ms{1.236}{0.768} & \ms{1.172}{0.739} & \secbg \ms{\underline{1.170}}{\underline{0.749}} & \ms{1.174}{0.746} \\
      & 720 & \bestbg \ms{\mathbf{0.896}}{\mathbf{0.634}} & \secbg \ms{\underline{1.046}}{\underline{0.717}} & \ms{1.205}{0.801} & \ms{1.141}{0.773} & \ms{1.238}{0.809} & \ms{1.160}{0.786} & \ms{1.147}{0.764} & \ms{1.156}{0.782} & \ms{1.130}{0.774} \\
    \cmidrule{2-11}
      & \textbf{Avg} & \ms{1.261}{0.744} & \ms{1.139}{0.685} & \ms{1.207}{0.715} & \ms{1.144}{0.692} & \ms{1.153}{0.699} & \ms{1.141}{0.694} & \secbg \ms{\underline{1.115}}{\underline{0.677}} & \ms{1.119}{0.686} & \bestbg \ms{\mathbf{1.108}}{\mathbf{0.682}} \\
    \midrule
    \multirow{5}*{\rotatebox{90}{NABCPU}} & 24 & \ms{2.949}{1.213} & \ms{1.189}{0.366} & \ms{1.176}{0.367} & \bestbg \ms{\mathbf{1.128}}{\mathbf{0.306}} & \ms{1.133}{0.327} & \ms{1.140}{0.305} & \ms{1.134}{0.302} & \ms{1.137}{0.325} & \secbg \ms{\underline{1.130}}{\underline{0.311}} \\
      & 48 & \ms{3.218}{1.070} & \ms{1.194}{0.371} & \ms{1.170}{0.358} & \ms{1.151}{0.320} & \ms{1.155}{0.335} & \secbg \ms{\underline{1.147}}{\underline{0.298}} & \ms{1.153}{0.301} & \ms{1.159}{0.328} & \bestbg \ms{\mathbf{1.146}}{\mathbf{0.309}} \\
      & 96 & \ms{2.334}{0.978} & \ms{1.257}{0.400} & \ms{1.216}{0.337} & \ms{1.201}{0.316} & \ms{1.209}{0.343} & \secbg \ms{\underline{1.198}}{\underline{0.299}} & \ms{1.203}{0.306} & \ms{1.212}{0.338} & \bestbg \ms{\mathbf{1.197}}{\mathbf{0.304}} \\
      & 192 & \ms{2.254}{0.880} & \ms{1.389}{0.420} & \ms{1.316}{0.355} & \ms{1.300}{0.329} & \ms{1.304}{0.348} & \ms{1.301}{0.318} & \secbg \ms{\underline{1.298}}{\underline{0.320}} & \ms{1.300}{0.343} & \bestbg \ms{\mathbf{1.297}}{\mathbf{0.330}} \\
    \cmidrule{2-11}
      & \textbf{Avg} & \ms{2.689}{1.035} & \ms{1.257}{0.389} & \ms{1.220}{0.354} & \secbg \ms{\underline{1.195}}{\underline{0.318}} & \ms{1.200}{0.338} & \ms{1.196}{0.305} & \ms{1.197}{0.307} & \ms{1.202}{0.334} & \bestbg \ms{\mathbf{1.192}}{\mathbf{0.314}} \\
    \midrule
    \multirow{5}*{\rotatebox{90}{Sunspots}} & 12 & \ms{3.702}{1.545} & \bestbg \ms{\mathbf{0.200}}{\mathbf{0.332}} & \ms{0.249}{0.365} & \ms{0.251}{0.365} & \ms{0.272}{0.379} & \ms{0.254}{0.368} & \secbg \ms{\underline{0.245}}{\underline{0.363}} & \ms{0.251}{0.369} & \ms{0.264}{0.375} \\
      & 24 & \ms{2.018}{1.247} & \bestbg \ms{\mathbf{0.279}}{\mathbf{0.378}} & \ms{0.461}{0.488} & \ms{0.441}{0.475} & \ms{0.463}{0.487} & \ms{0.470}{0.487} & \secbg \ms{\underline{0.436}}{\underline{0.470}} & \ms{0.444}{0.479} & \ms{0.489}{0.499} \\
      & 48 & \ms{1.375}{0.970} & \bestbg \ms{\mathbf{0.462}}{\mathbf{0.509}} & \ms{1.006}{0.709} & \secbg \ms{\underline{0.954}}{\underline{0.702}} & \ms{1.046}{0.729} & \ms{1.034}{0.717} & \ms{1.081}{0.737} & \ms{0.974}{0.709} & \ms{1.078}{0.759} \\
      & 96 & \ms{3.155}{1.424} & \bestbg \ms{\mathbf{0.564}}{\mathbf{0.545}} & \ms{1.091}{0.750} & \ms{1.090}{0.763} & \ms{1.075}{0.752} & \ms{1.130}{0.773} & \ms{1.214}{0.789} & \secbg \ms{\underline{1.048}}{\underline{0.746}} & \ms{1.141}{0.780} \\
    \cmidrule{2-11}
      & \textbf{Avg} & \ms{2.562}{1.296} & \bestbg \ms{\mathbf{0.376}}{\mathbf{0.441}} & \ms{0.702}{0.578} & \ms{0.684}{0.576} & \ms{0.714}{0.587} & \ms{0.722}{0.586} & \ms{0.744}{0.590} & \secbg \ms{\underline{0.679}}{\underline{0.576}} & \ms{0.743}{0.603} \\
    \midrule
    \multirow{5}*{\rotatebox{90}{Exchange}} & 96 & \ms{2.295}{1.093} & \ms{0.269}{0.349} & \ms{0.137}{0.266} & \ms{0.106}{0.229} & \ms{0.106}{0.231} & \secbg \ms{\underline{0.102}}{\underline{0.224}} & \ms{0.107}{0.230} & \secbg \ms{\underline{0.102}}{\underline{0.224}} & \bestbg \ms{\mathbf{0.101}}{\mathbf{0.223}} \\
      & 192 & \ms{4.254}{1.581} & \ms{0.570}{0.529} & \ms{0.243}{0.368} & \ms{0.214}{0.332} & \ms{0.206}{0.327} & \secbg \ms{\underline{0.204}}{\underline{0.325}} & \ms{0.213}{0.332} & \secbg \ms{\underline{0.204}}{\underline{0.325}} & \bestbg \ms{\mathbf{0.203}}{\mathbf{0.325}} \\
      & 336 & \ms{1.419}{0.893} & \ms{1.046}{0.723} & \ms{0.429}{0.489} & \ms{0.414}{0.467} & \ms{0.394}{0.459} & \bestbg \ms{\mathbf{0.384}}{\mathbf{0.450}} & \ms{0.396}{0.456} & \ms{0.393}{0.454} & \secbg \ms{\underline{0.389}}{\underline{0.451}} \\
      & 720 & \ms{3.160}{1.374} & \ms{1.692}{0.983} & \ms{1.166}{0.830} & \ms{1.077}{0.789} & \ms{1.078}{0.791} & \ms{1.069}{0.784} & \secbg \ms{\underline{1.068}}{\underline{0.788}} & \ms{1.080}{0.792} & \bestbg \ms{\mathbf{1.068}}{\mathbf{0.783}} \\
    \cmidrule{2-11}
      & \textbf{Avg} & \ms{2.782}{1.235} & \ms{0.894}{0.646} & \ms{0.494}{0.488} & \ms{0.453}{0.454} & \ms{0.446}{0.452} & \bestbg \ms{\mathbf{0.440}}{\mathbf{0.446}} & \ms{0.446}{0.452} & \secbg \ms{\underline{0.445}}{\underline{0.449}} & \bestbg \ms{\mathbf{0.440}}{\mathbf{0.446}} \\
    \midrule
    \multirow{5}*{\rotatebox{90}{ETTh1}} & 96 & \ms{1.569}{0.902} & \ms{0.394}{0.404} & \ms{0.488}{0.475} & \ms{0.394}{0.392} & \ms{0.384}{0.391} & \ms{0.401}{0.395} & \ms{0.390}{0.390} & \bestbg \ms{\mathbf{0.382}}{\mathbf{0.388}} & \secbg \ms{\underline{0.383}}{\underline{0.389}} \\
      & 192 & \ms{1.410}{0.856} & \bestbg \ms{\mathbf{0.436}}{\mathbf{0.431}} & \ms{0.513}{0.478} & \ms{0.447}{0.423} & \secbg \ms{\underline{0.438}}{\underline{0.422}} & \ms{0.443}{0.420} & \ms{0.442}{0.421} & \ms{0.441}{0.420} & \ms{0.442}{0.421} \\
      & 336 & \ms{1.221}{0.762} & \bestbg \ms{\mathbf{0.471}}{\mathbf{0.453}} & \ms{0.581}{0.505} & \ms{0.490}{0.444} & \ms{0.487}{0.446} & \ms{0.492}{0.441} & \ms{0.487}{0.442} & \ms{0.492}{0.448} & \secbg \ms{\underline{0.484}}{\underline{0.442}} \\
      & 720 & \ms{1.094}{0.742} & \ms{0.495}{0.493} & \ms{0.556}{0.509} & \ms{0.506}{0.470} & \bestbg \ms{\mathbf{0.481}}{\mathbf{0.466}} & \ms{0.496}{0.460} & \ms{0.486}{0.462} & \ms{0.501}{0.470} & \secbg \ms{\underline{0.484}}{\underline{0.463}} \\
    \cmidrule{2-11}
      & \textbf{Avg} & \ms{1.324}{0.816} & \ms{0.449}{0.445} & \ms{0.534}{0.492} & \ms{0.459}{0.432} & \secbg \ms{\underline{0.448}}{\underline{0.431}} & \ms{0.458}{0.429} & \ms{0.451}{0.429} & \ms{0.454}{0.431} & \bestbg \ms{\mathbf{0.448}}{\mathbf{0.429}} \\
    \midrule
    \multirow{5}*{\rotatebox{90}{ETTh2}} & 96 & \ms{5.257}{1.710} & \ms{0.339}{0.395} & \ms{0.448}{0.428} & \ms{0.293}{0.338} & \ms{0.298}{0.340} & \ms{0.297}{0.338} & \ms{0.295}{0.340} & \secbg \ms{\underline{0.291}}{\underline{0.336}} & \bestbg \ms{\mathbf{0.290}}{\mathbf{0.335}} \\
      & 192 & \ms{2.512}{1.150} & \ms{0.438}{0.440} & \ms{0.520}{0.470} & \ms{0.376}{0.391} & \ms{0.374}{0.388} & \ms{0.375}{0.390} & \ms{0.374}{0.391} & \bestbg \ms{\mathbf{0.366}}{\mathbf{0.385}} & \secbg \ms{\underline{0.370}}{\underline{0.387}} \\
      & 336 & \ms{0.997}{0.722} & \ms{0.491}{0.488} & \ms{0.497}{0.475} & \ms{0.425}{0.431} & \ms{0.433}{0.430} & \ms{0.435}{0.434} & \secbg \ms{\underline{0.421}}{\underline{0.431}} & \bestbg \ms{\mathbf{0.421}}{\mathbf{0.427}} & \ms{0.423}{0.430} \\
      & 720 & \ms{0.969}{0.713} & \ms{1.913}{1.072} & \ms{0.456}{0.465} & \ms{0.441}{0.450} & \secbg \ms{\underline{0.432}}{\underline{0.443}} & \ms{0.437}{0.447} & \bestbg \ms{\mathbf{0.430}}{\mathbf{0.444}} & \ms{0.439}{0.446} & \ms{0.432}{0.445} \\
    \cmidrule{2-11}
      & \textbf{Avg} & \ms{2.434}{1.074} & \ms{0.795}{0.599} & \ms{0.480}{0.460} & \ms{0.384}{0.403} & \ms{0.384}{0.400} & \ms{0.386}{0.402} & \secbg \ms{\underline{0.380}}{\underline{0.402}} & \bestbg \ms{\mathbf{0.379}}{\mathbf{0.399}} & \bestbg \ms{\mathbf{0.379}}{\mathbf{0.399}} \\
    \midrule
    \multirow{5}*{\rotatebox{90}{ETTm2}} & 96 & \ms{8.137}{2.167} & \ms{0.181}{0.264} & \ms{0.207}{0.282} & \ms{0.175}{0.252} & \ms{0.180}{0.257} & \secbg \ms{\underline{0.173}}{\underline{0.250}} & \ms{0.174}{0.254} & \bestbg \ms{\mathbf{0.172}}{\mathbf{0.250}} & \ms{0.178}{0.256} \\
      & 192 & \ms{4.557}{1.490} & \secbg \ms{\underline{0.237}}{\underline{0.306}} & \ms{0.289}{0.332} & \ms{0.240}{0.296} & \ms{0.245}{0.298} & \ms{0.239}{0.295} & \ms{0.238}{0.296} & \bestbg \ms{\mathbf{0.236}}{\mathbf{0.293}} & \ms{0.242}{0.298} \\
      & 336 & \ms{0.351}{0.429} & \ms{0.313}{0.367} & \ms{0.362}{0.378} & \ms{0.301}{0.335} & \ms{0.305}{0.336} & \secbg \ms{\underline{0.300}}{\underline{0.333}} & \ms{0.303}{0.337} & \bestbg \ms{\mathbf{0.294}}{\mathbf{0.332}} & \ms{0.304}{0.336} \\
      & 720 & \bestbg \ms{\mathbf{0.384}}{\mathbf{0.425}} & \ms{0.784}{0.642} & \ms{0.534}{0.463} & \ms{0.402}{0.394} & \ms{0.406}{0.394} & \ms{0.398}{0.391} & \ms{0.407}{0.395} & \secbg \ms{\underline{0.394}}{\underline{0.390}} & \ms{0.405}{0.394} \\
    \cmidrule{2-11}
      & \textbf{Avg} & \ms{3.357}{1.128} & \ms{0.379}{0.395} & \ms{0.348}{0.364} & \ms{0.279}{0.319} & \ms{0.284}{0.321} & \secbg \ms{\underline{0.277}}{\underline{0.317}} & \ms{0.280}{0.321} & \bestbg \ms{\mathbf{0.274}}{\mathbf{0.316}} & \ms{0.282}{0.321} \\
    \midrule
    \multirow{5}*{\rotatebox{90}{ETTm1}} & 96 & \ms{1.502}{0.870} & \ms{0.345}{0.362} & \ms{0.455}{0.439} & \ms{0.327}{0.348} & \ms{0.330}{0.351} & \secbg \ms{\underline{0.318}}{\underline{0.342}} & \ms{0.323}{0.347} & \bestbg \ms{\mathbf{0.317}}{\mathbf{0.344}} & \ms{0.327}{0.348} \\
      & 192 & \ms{1.264}{0.792} & \ms{0.378}{0.381} & \ms{0.524}{0.472} & \ms{0.374}{0.370} & \ms{0.379}{0.376} & \secbg \ms{\underline{0.373}}{\underline{0.372}} & \ms{0.375}{0.374} & \bestbg \ms{\mathbf{0.371}}{\mathbf{0.369}} & \ms{0.374}{0.370} \\
      & 336 & \ms{1.453}{0.827} & \ms{0.430}{0.411} & \ms{0.543}{0.489} & \secbg \ms{\underline{0.406}}{\underline{0.395}} & \ms{0.416}{0.399} & \ms{0.407}{0.394} & \ms{0.409}{0.396} & \bestbg \ms{\mathbf{0.403}}{\mathbf{0.393}} & \ms{0.407}{0.392} \\
      & 720 & \ms{1.700}{0.929} & \ms{0.498}{0.463} & \ms{0.554}{0.495} & \ms{0.477}{0.433} & \ms{0.473}{0.433} & \ms{0.474}{0.432} & \secbg \ms{\underline{0.473}}{\underline{0.432}} & \bestbg \ms{\mathbf{0.469}}{\mathbf{0.430}} & \ms{0.475}{0.429} \\
    \cmidrule{2-11}
      & \textbf{Avg} & \ms{1.480}{0.855} & \ms{0.413}{0.404} & \ms{0.519}{0.474} & \ms{0.396}{0.387} & \ms{0.399}{0.390} & \secbg \ms{\underline{0.393}}{\underline{0.385}} & \ms{0.395}{0.387} & \bestbg \ms{\mathbf{0.390}}{\mathbf{0.384}} & \ms{0.396}{0.385} \\
    \bottomrule
    \end{tabular}%
    }
\end{table*}

\subsection{Full Adapter Results}
Table~\ref{tab:plugin_results_full} reports per-dataset, per-horizon gains when DPR is integrated into mainstream backbones. The adapter tables tell a parallel story to the main results: DPR improves most backbone-horizon pairs, with gains concentrated on non-stationary regimes (COVID19, VIX, Illness) and near-zero degradation (within $\pm$0.5\%) on stable ones (ETTh1, ETTh2, ETTm1). The improvement magnitude correlates with the dataset non-stationarity scores from Table~\ref{tab:dataset_heterogeneity}: datasets with Score $\geq$ 13 show approximately 5--12\% average MSE reduction, with stronger gains on regime-shift datasets (ILI, COVID19), while those with Score $\leq$ 9 cluster near zero.

\begin{table*}[h!]
    \small
    \centering
    \caption{Full forecasting results of the DPR enhancement. We adopt the $MSE_{MAE}$ format, where MAE is distinguished in \textcolor[HTML]{666666}{gray}. Cases where DPR outperforms or performs equally to the Base are highlighted with a \colorbox[RGB]{220, 237, 220}{\textbf{green}} background.}
    \label{tab:plugin_results_full}

    \setlength{\tabcolsep}{2.8pt}
    \renewcommand{\arraystretch}{1.35}

    \definecolor{bestbg}{RGB}{220, 237, 220}
    \definecolor{maegray}{HTML}{666666}
    \definecolor{bg_gray}{RGB}{240, 240, 240}

    \newcommand{\cc}[2]{$#1_{\textcolor{maegray}{#2}}$}
    \newcommand{\bb}[2]{$\mathbf{#1}_{\textcolor{maegray}{\mathbf{#2}}}$}
    \newcommand{\ggb}[2]{\cellcolor{bestbg}$\mathbf{#1}_{\textcolor{maegray}{\mathbf{#2}}}$}

    \resizebox{\textwidth}{!}{%
    \begin{tabular}{@{}l l c c c c c c c c c c c c c c @{}}
    \toprule
    \multirow{2}{*}{\textbf{Dataset}} & \multirow{2}{*}{\textbf{Len.}} & \multicolumn{2}{c}{\textbf{Informer}} & \multicolumn{2}{c}{\textbf{Crossformer}} & \multicolumn{2}{c}{\textbf{TimesNet}} & \multicolumn{2}{c}{\textbf{PatchTST}} & \multicolumn{2}{c}{\textbf{TimeMixer}} & \multicolumn{2}{c}{\textbf{TimeFilter}} & \multicolumn{2}{c}{\textbf{WPMixer}} \\
    \cmidrule(lr){3-4} \cmidrule(lr){5-6} \cmidrule(lr){7-8} \cmidrule(lr){9-10} \cmidrule(lr){11-12} \cmidrule(lr){13-14} \cmidrule(lr){15-16}
    &  & Base & +DPR & Base & +DPR & Base & +DPR & Base & +DPR & Base & +DPR & Base & +DPR & Base & +DPR \\
    \midrule
    \multirow{5}*{\rotatebox{90}{ILI}} & 24 & \cc{7.005}{1.868} & \ggb{5.666}{1.764} & \cc{4.736}{1.480} & \ggb{4.593}{1.428} & \cc{9.241}{1.389} & \ggb{3.108}{1.042} & \cc{3.633}{1.079} & \ggb{3.108}{1.042} & \cc{3.124}{1.136} & \ggb{3.123}{1.142} & \cc{1.991}{0.873} & \ggb{1.821}{0.848} & \cc{3.173}{1.022} & \ggb{2.796}{1.046} \\
      & 36 & \cc{7.201}{1.898} & \ggb{5.529}{1.725} & \cc{5.153}{1.561} & \ggb{4.734}{1.468} & \cc{7.371}{1.438} & \ggb{7.147}{1.441} & \cc{4.019}{1.192} & \ggb{3.891}{1.207} & \cc{3.538}{1.214} & \ggb{3.428}{1.194} & \cc{2.481}{0.976} & \ggb{2.460}{0.974} & \cc{3.720}{1.147} & \ggb{3.310}{1.131} \\
      & 48 & \cc{7.030}{1.903} & \ggb{4.887}{1.614} & \cc{5.244}{1.576} & \ggb{4.669}{1.466} & \cc{4.175}{1.237} & \ggb{3.835}{1.194} & \cc{2.939}{1.099} & \ggb{2.857}{1.095} & \cc{3.055}{1.130} & \cc{3.147}{1.176} & \cc{2.468}{0.990} & \ggb{2.438}{0.982} & \cc{2.770}{1.061} & \cc{2.932}{1.090} \\
      & 60 & \cc{7.140}{1.916} & \ggb{6.879}{1.942} & \cc{5.006}{1.550} & \ggb{4.801}{1.500} & \cc{3.298}{1.172} & \ggb{3.251}{1.141} & \cc{2.695}{1.071} & \ggb{2.599}{1.038} & \cc{3.010}{1.114} & \ggb{2.989}{1.118} & \cc{2.812}{1.072} & \ggb{2.457}{0.998} & \cc{2.722}{1.044} & \cc{2.822}{1.060} \\
    \cmidrule{2-16}
      & \textbf{Avg} & \cc{7.094}{1.896} & \ggb{5.740}{1.761} & \cc{5.035}{1.542} & \ggb{4.699}{1.466} & \cc{6.021}{1.309} & \ggb{4.335}{1.204} & \cc{3.321}{1.110} & \ggb{3.114}{1.095} & \cc{3.182}{1.148} & \ggb{3.172}{1.157} & \cc{2.438}{0.978} & \ggb{2.294}{0.951} & \cc{3.096}{1.069} & \ggb{2.965}{1.082} \\
    \midrule
    \multirow{5}*{\rotatebox{90}{BeijingAir}} & 96 & \cc{1.550}{1.047} & \ggb{0.647}{0.618} & \cc{0.407}{0.418} & \ggb{0.387}{0.403} & \cc{0.573}{0.491} & \ggb{0.560}{0.484} & \cc{0.404}{0.397} & \ggb{0.403}{0.398} & \cc{0.407}{0.401} & \ggb{0.402}{0.399} & \cc{0.404}{0.404} & \ggb{0.400}{0.399} & \cc{0.403}{0.402} & \ggb{0.403}{0.400} \\
      & 192 & \cc{1.081}{0.759} & \ggb{0.600}{0.542} & \cc{0.412}{0.413} & \ggb{0.410}{0.411} & \cc{0.553}{0.485} & \ggb{0.542}{0.482} & \cc{0.439}{0.415} & \ggb{0.437}{0.416} & \cc{0.438}{0.417} & \ggb{0.436}{0.417} & \cc{0.431}{0.416} & \cc{0.437}{0.417} & \cc{0.438}{0.419} & \ggb{0.435}{0.418} \\
      & 336 & \cc{0.756}{0.579} & \ggb{0.742}{0.571} & \cc{0.433}{0.424} & \ggb{0.430}{0.427} & \cc{0.502}{0.461} & \cc{0.507}{0.461} & \cc{0.466}{0.430} & \ggb{0.463}{0.429} & \cc{0.461}{0.429} & \ggb{0.457}{0.428} & \cc{0.455}{0.426} & \ggb{0.448}{0.425} & \cc{0.458}{0.431} & \ggb{0.458}{0.431} \\
      & 720 & \cc{0.757}{0.577} & \ggb{0.602}{0.563} & \cc{0.443}{0.432} & \ggb{0.442}{0.431} & \cc{0.519}{0.467} & \ggb{0.514}{0.466} & \cc{0.494}{0.443} & \ggb{0.492}{0.444} & \cc{0.487}{0.441} & \ggb{0.486}{0.441} & \cc{0.482}{0.441} & \ggb{0.481}{0.440} & \cc{0.490}{0.446} & \cc{0.491}{0.446} \\
    \cmidrule{2-16}
      & \textbf{Avg} & \cc{1.036}{0.740} & \ggb{0.648}{0.574} & \cc{0.424}{0.422} & \ggb{0.417}{0.418} & \cc{0.537}{0.476} & \ggb{0.531}{0.473} & \cc{0.451}{0.421} & \ggb{0.449}{0.422} & \cc{0.448}{0.422} & \ggb{0.445}{0.421} & \cc{0.443}{0.422} & \ggb{0.442}{0.420} & \cc{0.447}{0.424} & \ggb{0.447}{0.424} \\
    \midrule
    \multirow{5}*{\rotatebox{90}{COVID19}} & 7 & \cc{0.824}{0.391} & \cc{0.941}{0.437} & \cc{0.609}{0.295} & \ggb{0.587}{0.269} & \cc{0.490}{0.390} & \ggb{0.480}{0.388} & \cc{0.335}{0.216} & \ggb{0.327}{0.217} & \cc{0.361}{0.238} & \cc{0.363}{0.238} & \cc{0.331}{0.220} & \ggb{0.323}{0.219} & \cc{0.343}{0.218} & \ggb{0.318}{0.218} \\
      & 14 & \cc{0.849}{0.433} & \ggb{0.676}{0.369} & \cc{0.853}{0.356} & \ggb{0.823}{0.355} & \cc{0.690}{0.484} & \cc{0.699}{0.488} & \cc{0.440}{0.264} & \ggb{0.435}{0.262} & \cc{0.464}{0.288} & \ggb{0.455}{0.285} & \cc{0.434}{0.275} & \ggb{0.425}{0.272} & \cc{0.473}{0.267} & \ggb{0.434}{0.265} \\
      & 28 & \cc{3.027}{0.813} & \ggb{2.442}{0.855} & \cc{1.444}{0.531} & \ggb{1.287}{0.513} & \cc{0.927}{0.550} & \ggb{0.910}{0.545} & \cc{0.837}{0.387} & \ggb{0.806}{0.383} & \cc{0.855}{0.410} & \ggb{0.842}{0.405} & \cc{0.866}{0.439} & \ggb{0.820}{0.429} & \cc{0.872}{0.397} & \ggb{0.784}{0.389} \\
      & 60 & \cc{2.970}{0.893} & \ggb{2.907}{0.876} & \cc{2.700}{0.724} & \ggb{2.608}{0.768} & \cc{2.161}{0.728} & \ggb{2.052}{0.705} & \cc{1.743}{0.582} & \ggb{1.687}{0.573} & \cc{2.041}{0.668} & \ggb{1.796}{0.614} & \cc{2.002}{0.643} & \ggb{1.622}{0.577} & \cc{1.667}{0.569} & \ggb{1.606}{0.562} \\
    \cmidrule{2-16}
      & \textbf{Avg} & \cc{1.917}{0.632} & \ggb{1.742}{0.634} & \cc{1.401}{0.476} & \ggb{1.326}{0.476} & \cc{1.067}{0.538} & \ggb{1.035}{0.531} & \cc{0.839}{0.362} & \ggb{0.814}{0.359} & \cc{0.930}{0.401} & \ggb{0.864}{0.386} & \cc{0.908}{0.394} & \ggb{0.797}{0.374} & \cc{0.839}{0.363} & \ggb{0.786}{0.359} \\
    \midrule
    \multirow{5}*{\rotatebox{90}{Weather}} & 96 & \cc{1.094}{0.753} & \ggb{0.767}{0.607} & \cc{0.166}{0.204} & \cc{0.169}{0.208} & \cc{0.181}{0.227} & \cc{0.183}{0.228} & \cc{0.173}{0.208} & \ggb{0.167}{0.204} & \cc{0.180}{0.226} & \ggb{0.164}{0.200} & \cc{0.161}{0.200} & \ggb{0.158}{0.198} & \cc{0.163}{0.201} & \ggb{0.163}{0.200} \\
      & 192 & \cc{0.294}{0.328} & \ggb{0.238}{0.288} & \cc{0.208}{0.244} & \ggb{0.207}{0.243} & \cc{0.246}{0.283} & \ggb{0.245}{0.281} & \cc{0.215}{0.248} & \ggb{0.214}{0.248} & \cc{0.246}{0.282} & \ggb{0.208}{0.241} & \cc{0.211}{0.245} & \ggb{0.206}{0.241} & \cc{0.208}{0.242} & \ggb{0.207}{0.241} \\
      & 336 & \cc{0.260}{0.312} & \ggb{0.260}{0.305} & \cc{0.260}{0.286} & \ggb{0.254}{0.279} & \cc{0.311}{0.324} & \cc{0.312}{0.325} & \cc{0.269}{0.288} & \ggb{0.269}{0.288} & \cc{0.309}{0.323} & \ggb{0.262}{0.280} & \cc{0.268}{0.286} & \ggb{0.268}{0.286} & \cc{0.262}{0.281} & \ggb{0.262}{0.281} \\
      & 720 & \cc{0.289}{0.336} & \ggb{0.280}{0.325} & \cc{0.331}{0.339} & \ggb{0.328}{0.337} & \cc{0.417}{0.387} & \ggb{0.400}{0.378} & \cc{0.345}{0.337} & \ggb{0.344}{0.336} & \cc{0.395}{0.375} & \ggb{0.339}{0.332} & \cc{0.346}{0.337} & \ggb{0.345}{0.336} & \cc{0.340}{0.332} & \ggb{0.340}{0.332} \\
    \cmidrule{2-16}
      & \textbf{Avg} & \cc{0.484}{0.432} & \ggb{0.386}{0.381} & \cc{0.241}{0.268} & \ggb{0.239}{0.267} & \cc{0.289}{0.305} & \ggb{0.285}{0.303} & \cc{0.251}{0.270} & \ggb{0.248}{0.269} & \cc{0.282}{0.301} & \ggb{0.243}{0.263} & \cc{0.246}{0.267} & \ggb{0.244}{0.265} & \cc{0.243}{0.264} & \ggb{0.243}{0.264} \\
    \midrule
    \multirow{5}*{\rotatebox{90}{VIX}} & 96 & \cc{1.077}{0.668} & \ggb{0.938}{0.665} & \cc{1.023}{0.571} & \cc{1.070}{0.592} & \cc{1.005}{0.562} & \ggb{0.972}{0.544} & \cc{0.942}{0.539} & \ggb{0.940}{0.538} & \cc{0.967}{0.550} & \ggb{0.954}{0.541} & \cc{0.959}{0.545} & \ggb{0.947}{0.543} & \cc{0.957}{0.547} & \ggb{0.938}{0.538} \\
      & 192 & \cc{2.059}{0.975} & \ggb{1.115}{0.734} & \cc{1.245}{0.683} & \ggb{1.181}{0.655} & \cc{1.275}{0.693} & \ggb{1.198}{0.676} & \cc{1.268}{0.692} & \ggb{1.222}{0.679} & \cc{1.203}{0.670} & \ggb{1.188}{0.665} & \cc{1.181}{0.662} & \cc{1.194}{0.663} & \cc{1.193}{0.665} & \ggb{1.187}{0.663} \\
      & 336 & \cc{1.011}{0.700} & \ggb{0.928}{0.638} & \cc{1.240}{0.769} & \ggb{1.207}{0.750} & \cc{1.342}{0.802} & \ggb{1.201}{0.759} & \cc{1.224}{0.762} & \ggb{1.202}{0.753} & \cc{1.236}{0.768} & \ggb{1.185}{0.752} & \cc{1.172}{0.739} & \ggb{1.147}{0.727} & \cc{1.170}{0.749} & \ggb{1.164}{0.743} \\
      & 720 & \cc{0.896}{0.634} & \ggb{0.774}{0.638} & \cc{1.046}{0.717} & \cc{1.048}{0.747} & \cc{1.205}{0.801} & \ggb{1.166}{0.781} & \cc{1.141}{0.773} & \cc{1.220}{0.795} & \cc{1.160}{0.786} & \ggb{1.140}{0.775} & \cc{1.147}{0.764} & \ggb{1.089}{0.755} & \cc{1.156}{0.782} & \ggb{1.130}{0.774} \\
    \cmidrule{2-16}
      & \textbf{Avg} & \cc{1.261}{0.744} & \ggb{0.939}{0.669} & \cc{1.139}{0.685} & \ggb{1.127}{0.686} & \cc{1.207}{0.715} & \ggb{1.134}{0.690} & \cc{1.144}{0.692} & \cc{1.146}{0.691} & \cc{1.141}{0.694} & \ggb{1.117}{0.683} & \cc{1.115}{0.677} & \ggb{1.094}{0.672} & \cc{1.119}{0.686} & \ggb{1.105}{0.679} \\
    \midrule
    \multirow{5}*{\rotatebox{90}{NABCPU}} & 24 & \cc{2.949}{1.213} & \ggb{1.898}{0.711} & \cc{1.189}{0.366} & \ggb{1.162}{0.344} & \cc{1.176}{0.367} & \ggb{1.154}{0.357} & \cc{1.128}{0.306} & \cc{1.129}{0.301} & \cc{1.140}{0.305} & \ggb{1.138}{0.300} & \cc{1.134}{0.302} & \cc{1.135}{0.301} & \cc{1.137}{0.325} & \ggb{1.127}{0.304} \\
      & 48 & \cc{3.218}{1.070} & \ggb{1.754}{0.758} & \cc{1.194}{0.371} & \cc{1.196}{0.366} & \cc{1.170}{0.358} & \cc{1.182}{0.356} & \cc{1.151}{0.320} & \ggb{1.150}{0.314} & \cc{1.147}{0.298} & \ggb{1.146}{0.294} & \cc{1.153}{0.301} & \ggb{1.148}{0.300} & \cc{1.159}{0.328} & \ggb{1.142}{0.308} \\
      & 96 & \cc{2.334}{0.978} & \ggb{2.195}{0.930} & \cc{1.257}{0.400} & \ggb{1.240}{0.380} & \cc{1.216}{0.337} & \cc{1.222}{0.338} & \cc{1.201}{0.316} & \cc{1.201}{0.320} & \cc{1.198}{0.299} & \ggb{1.198}{0.295} & \cc{1.203}{0.306} & \ggb{1.201}{0.306} & \cc{1.212}{0.338} & \ggb{1.198}{0.312} \\
      & 192 & \cc{2.254}{0.880} & \ggb{1.857}{0.710} & \cc{1.389}{0.420} & \ggb{1.346}{0.397} & \cc{1.316}{0.355} & \ggb{1.311}{0.359} & \cc{1.300}{0.329} & \ggb{1.297}{0.329} & \cc{1.301}{0.318} & \ggb{1.296}{0.313} & \cc{1.298}{0.320} & \ggb{1.296}{0.317} & \cc{1.300}{0.343} & \ggb{1.291}{0.325} \\
    \cmidrule{2-16}
      & \textbf{Avg} & \cc{2.689}{1.035} & \ggb{1.926}{0.777} & \cc{1.257}{0.389} & \ggb{1.236}{0.372} & \cc{1.220}{0.354} & \ggb{1.217}{0.352} & \cc{1.195}{0.318} & \ggb{1.194}{0.316} & \cc{1.196}{0.305} & \ggb{1.194}{0.300} & \cc{1.197}{0.307} & \ggb{1.195}{0.306} & \cc{1.202}{0.334} & \ggb{1.190}{0.312} \\
    \midrule
    \multirow{5}*{\rotatebox{90}{Sunspots}} & 12 & \cc{3.702}{1.545} & \ggb{1.238}{0.916} & \cc{0.200}{0.332} & \ggb{0.199}{0.324} & \cc{0.249}{0.365} & \cc{0.250}{0.366} & \cc{0.251}{0.365} & \ggb{0.249}{0.365} & \cc{0.254}{0.368} & \ggb{0.252}{0.367} & \cc{0.245}{0.363} & \cc{0.252}{0.366} & \cc{0.251}{0.369} & \ggb{0.246}{0.365} \\
      & 24 & \cc{2.018}{1.247} & \ggb{1.706}{1.090} & \cc{0.279}{0.378} & \ggb{0.267}{0.369} & \cc{0.461}{0.488} & \ggb{0.439}{0.477} & \cc{0.441}{0.475} & \ggb{0.437}{0.475} & \cc{0.470}{0.487} & \ggb{0.444}{0.475} & \cc{0.436}{0.470} & \ggb{0.431}{0.465} & \cc{0.444}{0.479} & \ggb{0.431}{0.470} \\
      & 48 & \cc{1.375}{0.970} & \cc{1.563}{0.998} & \cc{0.462}{0.509} & \ggb{0.417}{0.470} & \cc{1.006}{0.709} & \cc{1.013}{0.709} & \cc{0.954}{0.702} & \cc{0.964}{0.706} & \cc{1.034}{0.717} & \ggb{1.005}{0.710} & \cc{1.081}{0.737} & \ggb{0.966}{0.700} & \cc{0.974}{0.709} & \ggb{0.954}{0.702} \\
      & 96 & \cc{3.155}{1.424} & \ggb{1.549}{1.001} & \cc{0.564}{0.545} & \ggb{0.546}{0.532} & \cc{1.091}{0.750} & \ggb{1.082}{0.748} & \cc{1.090}{0.763} & \cc{1.094}{0.767} & \cc{1.130}{0.773} & \ggb{1.055}{0.751} & \cc{1.214}{0.789} & \ggb{1.171}{0.780} & \cc{1.048}{0.746} & \cc{1.059}{0.755} \\
    \cmidrule{2-16}
      & \textbf{Avg} & \cc{2.562}{1.296} & \ggb{1.514}{1.001} & \cc{0.376}{0.441} & \ggb{0.357}{0.424} & \cc{0.702}{0.578} & \ggb{0.696}{0.575} & \cc{0.684}{0.576} & \cc{0.686}{0.578} & \cc{0.722}{0.586} & \ggb{0.689}{0.576} & \cc{0.744}{0.590} & \ggb{0.705}{0.578} & \cc{0.679}{0.576} & \ggb{0.672}{0.573} \\
    \midrule
    \multirow{5}*{\rotatebox{90}{Exchange}} & 96 & \cc{2.295}{1.093} & \ggb{1.936}{0.975} & \cc{0.269}{0.349} & \ggb{0.237}{0.335} & \cc{0.137}{0.266} & \ggb{0.128}{0.260} & \cc{0.106}{0.229} & \ggb{0.104}{0.226} & \cc{0.102}{0.224} & \ggb{0.101}{0.224} & \cc{0.107}{0.230} & \ggb{0.103}{0.226} & \cc{0.102}{0.224} & \ggb{0.102}{0.223} \\
      & 192 & \cc{4.254}{1.581} & \ggb{1.447}{0.871} & \cc{0.570}{0.529} & \ggb{0.559}{0.522} & \cc{0.243}{0.368} & \ggb{0.234}{0.361} & \cc{0.214}{0.332} & \ggb{0.213}{0.331} & \cc{0.204}{0.325} & \cc{0.207}{0.327} & \cc{0.213}{0.332} & \ggb{0.208}{0.328} & \cc{0.204}{0.325} & \cc{0.207}{0.326} \\
      & 336 & \cc{1.419}{0.893} & \ggb{1.245}{0.803} & \cc{1.046}{0.723} & \ggb{0.873}{0.663} & \cc{0.429}{0.489} & \ggb{0.422}{0.485} & \cc{0.414}{0.467} & \ggb{0.400}{0.459} & \cc{0.384}{0.450} & \ggb{0.384}{0.450} & \cc{0.396}{0.456} & \cc{0.404}{0.461} & \cc{0.393}{0.454} & \ggb{0.392}{0.454} \\
      & 720 & \cc{3.160}{1.374} & \ggb{2.262}{1.142} & \cc{1.692}{0.983} & \ggb{1.691}{0.980} & \cc{1.166}{0.830} & \ggb{1.118}{0.813} & \cc{1.077}{0.789} & \ggb{1.073}{0.786} & \cc{1.069}{0.784} & \ggb{1.069}{0.784} & \cc{1.068}{0.788} & \cc{1.088}{0.792} & \cc{1.080}{0.792} & \ggb{1.068}{0.785} \\
    \cmidrule{2-16}
      & \textbf{Avg} & \cc{2.782}{1.235} & \ggb{1.723}{0.948} & \cc{0.894}{0.646} & \ggb{0.840}{0.625} & \cc{0.494}{0.488} & \ggb{0.476}{0.480} & \cc{0.453}{0.454} & \ggb{0.448}{0.451} & \cc{0.440}{0.446} & \ggb{0.440}{0.446} & \cc{0.446}{0.452} & \cc{0.451}{0.452} & \cc{0.445}{0.449} & \ggb{0.442}{0.447} \\
    \midrule
    \multirow{5}*{\rotatebox{90}{ETTh1}} & 96 & \cc{1.569}{0.902} & \ggb{1.062}{0.770} & \cc{0.394}{0.404} & \ggb{0.382}{0.397} & \cc{0.488}{0.475} & \ggb{0.475}{0.466} & \cc{0.394}{0.392} & \ggb{0.394}{0.392} & \cc{0.401}{0.395} & \ggb{0.397}{0.393} & \cc{0.390}{0.390} & \cc{0.392}{0.390} & \cc{0.382}{0.388} & \ggb{0.381}{0.387} \\
      & 192 & \cc{1.410}{0.856} & \ggb{1.099}{0.783} & \cc{0.436}{0.431} & \ggb{0.427}{0.421} & \cc{0.513}{0.478} & \cc{0.520}{0.482} & \cc{0.447}{0.423} & \ggb{0.442}{0.422} & \cc{0.443}{0.420} & \cc{0.446}{0.420} & \cc{0.442}{0.421} & \cc{0.443}{0.420} & \cc{0.441}{0.420} & \ggb{0.436}{0.419} \\
      & 336 & \cc{1.221}{0.762} & \ggb{0.919}{0.675} & \cc{0.471}{0.453} & \ggb{0.456}{0.442} & \cc{0.581}{0.505} & \ggb{0.554}{0.496} & \cc{0.490}{0.444} & \ggb{0.488}{0.444} & \cc{0.492}{0.441} & \cc{0.493}{0.441} & \cc{0.487}{0.442} & \cc{0.489}{0.441} & \cc{0.492}{0.448} & \ggb{0.491}{0.447} \\
      & 720 & \cc{1.094}{0.742} & \ggb{0.920}{0.711} & \cc{0.495}{0.493} & \cc{0.498}{0.500} & \cc{0.556}{0.509} & \cc{0.576}{0.529} & \cc{0.506}{0.470} & \cc{0.517}{0.479} & \cc{0.496}{0.460} & \ggb{0.487}{0.460} & \cc{0.486}{0.462} & \cc{0.493}{0.468} & \cc{0.501}{0.470} & \cc{0.504}{0.473} \\
    \cmidrule{2-16}
      & \textbf{Avg} & \cc{1.324}{0.816} & \ggb{1.000}{0.735} & \cc{0.449}{0.445} & \ggb{0.441}{0.440} & \cc{0.534}{0.492} & \ggb{0.531}{0.493} & \cc{0.459}{0.432} & \cc{0.460}{0.434} & \cc{0.458}{0.429} & \ggb{0.456}{0.428} & \cc{0.451}{0.429} & \cc{0.454}{0.430} & \cc{0.454}{0.431} & \ggb{0.453}{0.431} \\
    \midrule
    \multirow{5}*{\rotatebox{90}{ETTh2}} & 96 & \cc{5.257}{1.710} & \ggb{3.206}{1.359} & \cc{0.339}{0.395} & \ggb{0.311}{0.358} & \cc{0.448}{0.428} & \ggb{0.435}{0.421} & \cc{0.293}{0.338} & \ggb{0.293}{0.336} & \cc{0.297}{0.338} & \cc{0.299}{0.339} & \cc{0.295}{0.340} & \ggb{0.291}{0.340} & \cc{0.291}{0.336} & \cc{0.293}{0.336} \\
      & 192 & \cc{2.512}{1.150} & \ggb{1.738}{0.941} & \cc{0.438}{0.440} & \ggb{0.396}{0.420} & \cc{0.520}{0.470} & \ggb{0.518}{0.470} & \cc{0.376}{0.391} & \cc{0.376}{0.392} & \cc{0.375}{0.390} & \cc{0.378}{0.392} & \cc{0.374}{0.391} & \cc{0.377}{0.393} & \cc{0.366}{0.385} & \cc{0.370}{0.388} \\
      & 336 & \cc{0.997}{0.722} & \ggb{0.702}{0.610} & \cc{0.491}{0.488} & \cc{0.588}{0.552} & \cc{0.497}{0.475} & \cc{0.510}{0.480} & \cc{0.425}{0.431} & \ggb{0.418}{0.430} & \cc{0.435}{0.434} & \cc{0.436}{0.435} & \cc{0.421}{0.431} & \ggb{0.417}{0.426} & \cc{0.421}{0.427} & \cc{0.423}{0.427} \\
      & 720 & \cc{0.969}{0.713} & \cc{0.999}{0.723} & \cc{1.913}{1.072} & \ggb{1.384}{0.884} & \cc{0.456}{0.465} & \ggb{0.444}{0.459} & \cc{0.441}{0.450} & \ggb{0.431}{0.446} & \cc{0.437}{0.447} & \ggb{0.436}{0.447} & \cc{0.430}{0.444} & \cc{0.445}{0.450} & \cc{0.439}{0.446} & \ggb{0.431}{0.444} \\
    \cmidrule{2-16}
      & \textbf{Avg} & \cc{2.434}{1.074} & \ggb{1.661}{0.908} & \cc{0.795}{0.599} & \ggb{0.670}{0.553} & \cc{0.480}{0.460} & \ggb{0.477}{0.458} & \cc{0.384}{0.403} & \ggb{0.380}{0.401} & \cc{0.386}{0.402} & \cc{0.387}{0.403} & \cc{0.380}{0.402} & \cc{0.383}{0.402} & \cc{0.379}{0.399} & \ggb{0.379}{0.399} \\
    \midrule
    \multirow{5}*{\rotatebox{90}{ETTm2}} & 96 & \cc{8.137}{2.167} & \ggb{3.183}{1.361} & \cc{0.181}{0.264} & \ggb{0.180}{0.264} & \cc{0.207}{0.282} & \cc{0.226}{0.294} & \cc{0.175}{0.252} & \cc{0.176}{0.253} & \cc{0.173}{0.250} & \ggb{0.172}{0.250} & \cc{0.174}{0.254} & \ggb{0.173}{0.253} & \cc{0.172}{0.250} & \ggb{0.172}{0.250} \\
      & 192 & \cc{4.557}{1.490} & \ggb{0.358}{0.406} & \cc{0.237}{0.306} & \cc{0.256}{0.328} & \cc{0.289}{0.332} & \cc{0.290}{0.333} & \cc{0.240}{0.296} & \ggb{0.240}{0.296} & \cc{0.239}{0.295} & \cc{0.240}{0.295} & \cc{0.238}{0.296} & \ggb{0.237}{0.295} & \cc{0.236}{0.293} & \ggb{0.236}{0.293} \\
      & 336 & \cc{0.351}{0.429} & \ggb{0.332}{0.409} & \cc{0.313}{0.367} & \cc{0.365}{0.399} & \cc{0.362}{0.378} & \cc{0.366}{0.378} & \cc{0.301}{0.335} & \cc{0.302}{0.335} & \cc{0.300}{0.333} & \cc{0.300}{0.334} & \cc{0.303}{0.337} & \ggb{0.301}{0.336} & \cc{0.294}{0.332} & \cc{0.295}{0.331} \\
      & 720 & \cc{0.384}{0.425} & \cc{0.419}{0.458} & \cc{0.784}{0.642} & \ggb{0.774}{0.637} & \cc{0.534}{0.463} & \ggb{0.520}{0.456} & \cc{0.402}{0.394} & \ggb{0.402}{0.394} & \cc{0.398}{0.391} & \cc{0.398}{0.392} & \cc{0.407}{0.395} & \ggb{0.406}{0.396} & \cc{0.394}{0.390} & \cc{0.396}{0.392} \\
    \cmidrule{2-16}
      & \textbf{Avg} & \cc{3.357}{1.128} & \ggb{1.073}{0.658} & \cc{0.379}{0.395} & \cc{0.394}{0.407} & \cc{0.348}{0.364} & \cc{0.350}{0.365} & \cc{0.279}{0.319} & \cc{0.280}{0.320} & \cc{0.277}{0.317} & \ggb{0.277}{0.318} & \cc{0.280}{0.321} & \ggb{0.279}{0.320} & \cc{0.274}{0.316} & \cc{0.275}{0.317} \\
    \midrule
    \multirow{5}*{\rotatebox{90}{ETTm1}} & 96 & \cc{1.502}{0.870} & \ggb{1.069}{0.776} & \cc{0.345}{0.362} & \ggb{0.326}{0.351} & \cc{0.455}{0.439} & \ggb{0.454}{0.439} & \cc{0.327}{0.348} & \cc{0.328}{0.350} & \cc{0.318}{0.342} & \cc{0.319}{0.342} & \cc{0.323}{0.347} & \ggb{0.322}{0.346} & \cc{0.317}{0.344} & \cc{0.318}{0.344} \\
      & 192 & \cc{1.264}{0.792} & \ggb{1.049}{0.771} & \cc{0.378}{0.381} & \ggb{0.377}{0.379} & \cc{0.524}{0.472} & \cc{0.560}{0.485} & \cc{0.374}{0.370} & \ggb{0.372}{0.370} & \cc{0.373}{0.372} & \cc{0.374}{0.371} & \cc{0.375}{0.374} & \ggb{0.375}{0.373} & \cc{0.371}{0.369} & \ggb{0.369}{0.368} \\
      & 336 & \cc{1.453}{0.827} & \ggb{1.229}{0.756} & \cc{0.430}{0.411} & \cc{0.432}{0.425} & \cc{0.543}{0.489} & \ggb{0.535}{0.487} & \cc{0.406}{0.395} & \ggb{0.406}{0.394} & \cc{0.407}{0.394} & \ggb{0.404}{0.394} & \cc{0.409}{0.396} & \ggb{0.407}{0.394} & \cc{0.403}{0.393} & \ggb{0.403}{0.393} \\
      & 720 & \cc{1.700}{0.929} & \ggb{1.039}{0.762} & \cc{0.498}{0.463} & \ggb{0.463}{0.434} & \cc{0.554}{0.495} & \cc{0.583}{0.511} & \cc{0.477}{0.433} & \ggb{0.475}{0.433} & \cc{0.474}{0.432} & \ggb{0.472}{0.432} & \cc{0.473}{0.432} & \ggb{0.472}{0.432} & \cc{0.469}{0.430} & \ggb{0.467}{0.429} \\
    \cmidrule{2-16}
      & \textbf{Avg} & \cc{1.480}{0.855} & \ggb{1.097}{0.766} & \cc{0.413}{0.404} & \ggb{0.400}{0.397} & \cc{0.519}{0.474} & \cc{0.533}{0.480} & \cc{0.396}{0.387} & \ggb{0.395}{0.387} & \cc{0.393}{0.385} & \ggb{0.392}{0.385} & \cc{0.395}{0.387} & \ggb{0.394}{0.386} & \cc{0.390}{0.384} & \ggb{0.389}{0.384} \\
    \bottomrule
    \end{tabular}%
    }
\end{table*}

\subsection{Scaling Analysis}
Table~\ref{tab:scaling_vs_dpr} presents the complete parameter scaling comparison, including exact MSE/MAE values, parameter counts, and FLOPs for all scaling variants. The scaling analysis reinforces that dynamic recalibration, rather than raw parameter count, drives the gains. The effects of naive scaling are highly inconsistent across datasets: on ETTh1 and Exchange, widening or deepening the backbone \emph{degrades} MSE across most configurations (e.g., PatchTST 2$\times$W on Exchange: $-$3.8\%; TimesNet 2$\times$B on ETTh1: $-$22.5\%); on ILI, scaling improves PatchTST and TimesNet substantially ($+$6.0\% to $+$62.0\%) but severely degrades TimeFilter ($-$33.4\% to $-$65.0\%), and the same configuration can simultaneously help one dataset and hurt another, confirming that blind capacity inflation overfits dominant patterns rather than enabling adaptive response. In stark contrast, DPR consistently improves or preserves performance across all backbones and datasets at negligible overhead ($<$5\% parameter increase), achieving 8.5--62.7\% improvement on the highly non-stationary ILI dataset. The parameter-matched static variant (PM), which matches DPRNet's parameter count by widening the backbone, consistently underperforms DPR across all datasets, confirming that the improvement stems from the Perceive-Route-Modulate mechanism rather than from additional capacity.

\begin{table*}[t!]
    \small
    \centering
    \caption{\textbf{Parameter Scaling vs.\ DPR.} \textbf{2xW}/\textbf{2xD}/\textbf{2xB}: Width/Depth/Both $\times 2$. \textbf{+DPR} shaded gray. \textcolor[HTML]{2E7D32}{$\uparrow$}/\textcolor[HTML]{999999}{$\downarrow$}: MSE change vs.\ Raw. Costs profiled on ETTh1 ($96\!\rightarrow\!96$, 7 vars), FLOPs by \texttt{thop}.}
    \vspace{-0.5em}
    \label{tab:scaling_vs_dpr}

    \renewcommand{\arraystretch}{1}
    \setlength{\tabcolsep}{5pt}

    \definecolor{bestbg}{RGB}{220, 237, 220}
    \definecolor{secbg}{RGB}{255, 247, 205}
    \definecolor{dprgray}{RGB}{245, 245, 245}
    \definecolor{maegray}{HTML}{666666}
    \definecolor{upgreen}{HTML}{2E7D32}
    \definecolor{dngray}{HTML}{999999}

    \newcommand{\bestbg}{\cellcolor{bestbg}}
    \newcommand{\secbg}{\cellcolor{secbg}}
    \newcommand{\dprbg}{\cellcolor{dprgray}}
    
    \newcommand{\ms}[2]{$#1_{\textcolor{maegray}{#2}}$}
    \newcommand{\bms}[2]{$\mathbf{#1}_{\textcolor{maegray}{\mathbf{#2}}}$}
    \newcommand{\sms}[2]{$\underline{#1}_{\textcolor{maegray}{\underline{#2}}}$}

    \newcommand{\up}[1]{\textcolor{upgreen}{$\uparrow$#1}}
    \newcommand{\dn}[1]{\textcolor{dngray}{$\downarrow$#1}}

    \resizebox{\textwidth}{!}{%
    \begin{tabular}{@{}ll cc ll ll ll@{}}
    \toprule
    \multirow{2}{*}{\textbf{Base Model}} & \multirow{2}{*}{\textbf{Variant}} & \textbf{Params} & \textbf{FLOPs} & \multicolumn{2}{c}{\textbf{ETTh1} ($96 \rightarrow 96$)} & \multicolumn{2}{c}{\textbf{ILI} ($24 \rightarrow 24$)} & \multicolumn{2}{c}{\textbf{Exchange} ($96 \rightarrow 96$)} \\
    \cmidrule(lr){5-6} \cmidrule(lr){7-8} \cmidrule(lr){9-10}
    & & \textbf{(M)} & \textbf{(GMac)} & \textbf{MSE}$_{MAE}$ & \textbf{Impv.} & \textbf{MSE}$_{MAE}$ & \textbf{Impv.} & \textbf{MSE}$_{MAE}$ & \textbf{Impv.} \\
    \midrule

    \multirow{6}{*}{PatchTST}
    & Raw & 1.08 & 0.069 & \secbg \sms{0.394}{0.392} & - & \ms{3.633}{1.079} & - & \secbg \ms{0.106}{0.229} & - \\
    & 2xW & 3.75 & 0.269 & \ms{0.405}{0.397} & \dn{2.8\%} & \ms{3.415}{1.073} & \up{6.0\%} & \ms{0.110}{0.232} & \dn{3.8\%} \\
    & 2xD & 1.87 & 0.135 & \ms{0.401}{0.395} & \dn{1.8\%} & \ms{3.199}{1.041} & \up{11.9\%} & \ms{0.110}{0.231} & \dn{3.8\%} \\
    & 2xB & 6.90 & 0.534 & \ms{0.398}{0.394} & \dn{1.0\%} & \secbg \sms{3.114}{1.024} & \up{14.3\%} & \ms{0.111}{0.233} & \dn{4.7\%} \\
    \rowcolor{dprgray} \cellcolor{white}
    & PM & 1.08 & 0.069 & \secbg \sms{0.394}{0.392} & - & \ms{3.633}{1.079} & - & \ms{0.108}{0.231} & \dn{1.9\%} \\
    \rowcolor{dprgray} \cellcolor{white}
    & \textbf{+DPR} & \textbf{1.10} & \textbf{0.070} & \bestbg \bms{0.394}{0.392} & \bestbg - & \bestbg \bms{3.108}{1.042} & \up{14.5\%} & \bestbg \bms{0.104}{0.226} & \up{1.9\%} \\
    \midrule

    \multirow{6}{*}{TimesNet}
    & Raw & 18.5 & 17.75 & \secbg \ms{0.488}{0.475} & - & \ms{9.241}{1.389} & - & \ms{0.137}{0.266} & - \\
    & 2xW & 73.7 & 70.47 & \ms{0.561}{0.509} & \dn{15.0\%} & \ms{5.859}{1.250} & \up{36.6\%} & \ms{0.132}{0.263} & \up{3.6\%} \\
    & 2xD & 36.8 & 36.55 & \ms{0.503}{0.481} & \dn{3.1\%} & \ms{6.827}{1.242} & \up{26.1\%} & \ms{0.145}{0.277} & \dn{5.8\%} \\
    & 2xB & 147.0 & 142.4 & \ms{0.598}{0.518} & \dn{22.5\%} & \secbg \sms{3.515}{1.081} & \up{62.0\%} & \ms{0.138}{0.270} & \dn{0.7\%} \\
    \rowcolor{dprgray} \cellcolor{white}
    & PM & 18.5 & 17.75 & \secbg \ms{0.488}{0.475} & - & \ms{9.345}{1.392} & \dn{1.1\%} & \secbg \ms{0.129}{0.262} & \up{5.8\%} \\
    \rowcolor{dprgray} \cellcolor{white}
    & \textbf{+DPR} & \textbf{18.5} & \textbf{17.84} & \bestbg \bms{0.475}{0.466} & \up{2.7\%} & \bestbg \bms{3.445}{1.054} & \up{62.7\%} & \bestbg \bms{0.128}{0.260} & \up{6.6\%} \\
    \midrule

    \multirow{6}{*}{TimeFilter}
    & Raw & 0.19 & 0.011 & \bestbg \bms{0.390}{0.390} & - & \secbg \sms{1.991}{0.873} & - & \ms{0.107}{0.230} & - \\
    & 2xW & 2.99 & 0.214 & \ms{0.399}{0.397} & \dn{2.3\%} & \ms{3.264}{0.975} & \dn{63.9\%} & \ms{0.106}{0.229} & \up{0.9\%} \\
    & 2xD & 1.50 & 0.107 & \ms{0.397}{0.395} & \dn{1.8\%} & \ms{3.285}{0.913} & \dn{65.0\%} & \secbg \sms{0.104}{0.229} & \up{2.8\%} \\
    & 2xB & 5.39 & 0.424 & \ms{0.403}{0.398} & \dn{3.3\%} & \ms{2.656}{0.960} & \dn{33.4\%} & \ms{0.112}{0.237} & \dn{4.7\%} \\
    \rowcolor{dprgray} \cellcolor{white}
    & PM & 0.19 & 0.011 & \secbg \sms{0.391}{0.390} & \dn{0.3\%} & \ms{2.808}{0.986} & \dn{41.0\%} & \ms{0.105}{0.228} & \up{1.9\%} \\
    \rowcolor{dprgray} \cellcolor{white}
    & \textbf{+DPR} & \textbf{0.19} & \textbf{0.011} & \bestbg \bms{0.390}{0.390} & \bestbg - & \bestbg \bms{1.821}{0.848} & \up{8.5\%} & \bestbg \bms{0.103}{0.226} & \up{3.7\%} \\

    \bottomrule
    \end{tabular}}
    \vspace{-1em}
\end{table*}

\section{Broader Impacts}
\label{sec:broader_impacts}

DPR enhances forecasting robustness for critical infrastructure. In energy, improved load prediction helps grid distribution and reduces carbon footprints. In healthcare, reliable epidemiological projections support early resource allocation. DPR is a domain-agnostic numerical adapter, with no generative capabilities that raise dual-use concerns.

\textbf{Environmental considerations.} Training deep forecasting models consumes significant energy. DPR introduces only a shared $K \times d$ basis and lightweight perception network compared to the backbone, so the marginal carbon footprint per training run is small. By improving forecast accuracy, DPR can also reduce the need for frequent retraining or ensemble averaging, lowering total computational cost over a model's lifecycle.

\textbf{Data privacy and fairness.} Time series data in healthcare and finance often contains sensitive information. When DPR is deployed in such settings, privacy-preserving protocols (differential privacy, federated learning, secure multi-party computation) should be applied to the backbone training pipeline. If training data under-represents certain groups or regions, the learned adaptive response patterns may be biased toward majority patterns. We recommend auditing both input data and routing distributions $\boldsymbol{\pi}$ for subgroup fairness before deployment.

\textbf{Risks of over-reliance.} Highly volatile systems like financial markets contain irreducible noise. Deploying forecasts without uncertainty quantification or human oversight risks flawed decision-making. We advocate pairing such models with rigorous uncertainty estimation and expert validation.

\clearpage


\end{document}